\documentclass[conference]{IEEEtran}
\usepackage{times}

\usepackage[numbers]{natbib}
\usepackage{graphicx}   
\usepackage{multicol}
\usepackage{basicmath}
\usepackage{amsmath} 
\usepackage{amssymb}  
\usepackage{units}
\usepackage[bookmarks=true]{hyperref}
\usepackage[capitalise]{cleveref}
\usepackage{diagbox}
\usepackage{subcaption}
\usepackage{color}
\usepackage[normalem]{ulem}

\hypersetup{pdfinfo={
   Title={Dynamic On-Palm Manipulation via Controlled Sliding},
   Author={William Yang and Michael Posa},
   Subject={Robotics: Science and Systems 2024}
}}

\newcommand{\genAcc}{\ddot{q}}

\newcommand{\taskSpacePos}{y}

\newcommand{\taskSpaceVel}{\dot{y}}

\newcommand{\taskSpaceAccCmd}{\ddot{y}_{cmd}}
\newcommand{\taskSpaceAccErr}{(\ddot{y} - \taskSpaceAccCmd)}
\newcommand{\contactJacobian}{J_{\lambda}}

\begin{document}

\newcommand{\Revision}[1]{{#1}}
\renewcommand{\sout}[1]{\ignorespaces}

\title{Dynamic On-Palm Manipulation via Controlled Sliding}

\author{William Yang and Michael Posa\\
	University of Pennsylvania\\
	Email: \texttt{{yangwill,posa}@seas.upenn.edu}\\
\url{https://dynamic-controlled-sliding.github.io/}
}

\maketitle

\begin{abstract}
Non-prehensile manipulation enables fast interactions with objects by circumventing the need to grasp and ungrasp as well as handling objects that cannot be grasped through force closure.
Current approaches to non-prehensile manipulation focus on static contacts, avoiding the underactuation that comes with sliding.
However, the ability to control sliding contact, essentially removing the no-slip constraint, opens up new possibilities in dynamic manipulation.
In this paper, we explore a challenging dynamic non-prehensile manipulation task that requires the consideration of the full spectrum of hybrid contact modes.
We leverage recent methods in contact-implicit MPC to handle the multi-modal planning aspect of the task.
We demonstrate, with careful consideration of integration between the simple model used for MPC and the low-level tracking controller, how contact-implicit MPC can be adapted to dynamic tasks.
Surprisingly, despite the known inaccuracies of frictional rigid contact models, our method is able to react to these inaccuracies while still quickly performing the task.
Moreover, we do not use common aids such as reference trajectories or motion primitives, highlighting the generality of our approach.
To the best of our knowledge, this is the first application of contact-implicit MPC to a dynamic manipulation task in three dimensions.

\end{abstract}


\section{Introduction}
\label{sec:introduction}

Recent advancements in robot manipulation have demonstrated impressive dexterity \cite{chen2023visual} and generality \cite{curtis2022long} \cite{chi2023diffusion}.
However, these methods largely focus on slow tasks that can be viewed from a quasi-static perspective.
As robots are increasingly being asked to perform manipulation tasks in logistics applications such as warehouse robotics, speed becomes a key driving metric.
While there are plenty of examples of dynamic manipulation, the methods are often achieved using ad-hoc, task-specific solutions \cite{ruggiero2018nonprehensile} and demonstrated on systems with few degrees of freedom and few contacts.
The desire for a general control framework for contact-rich tasks has resulted in many formulations for contact-implicit model predictive control \cite{aydinoglu2023consensus} \cite{kurtz2023inverse} \cite{le2024fast} (MPC), which can automatically plan when and where to make and break contact and are reported to be fast enough for real-time control.

\begin{figure}[ht!]
	\centering
	\includegraphics[width=0.48\textwidth]{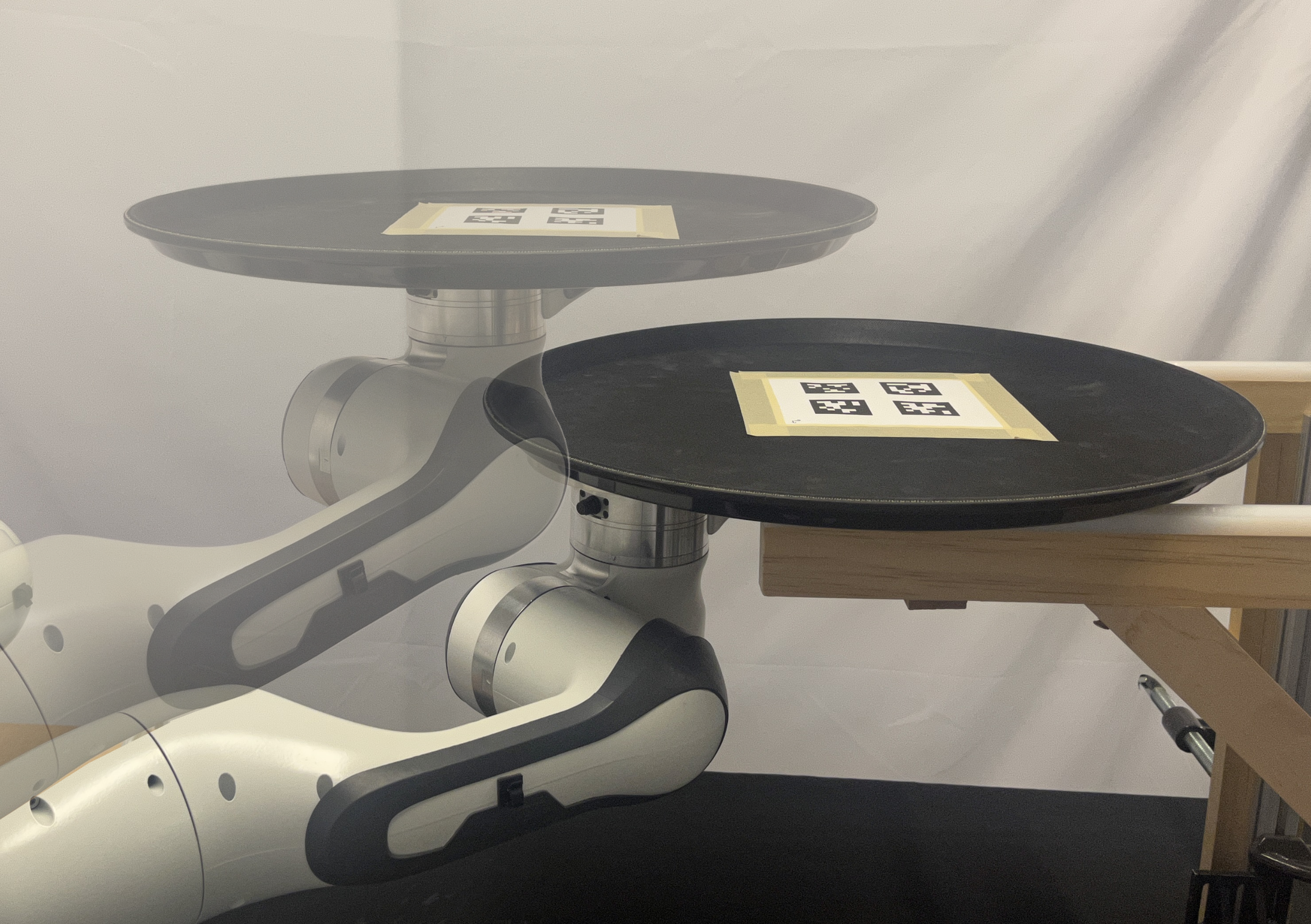}
	\includegraphics[height=6.7em]{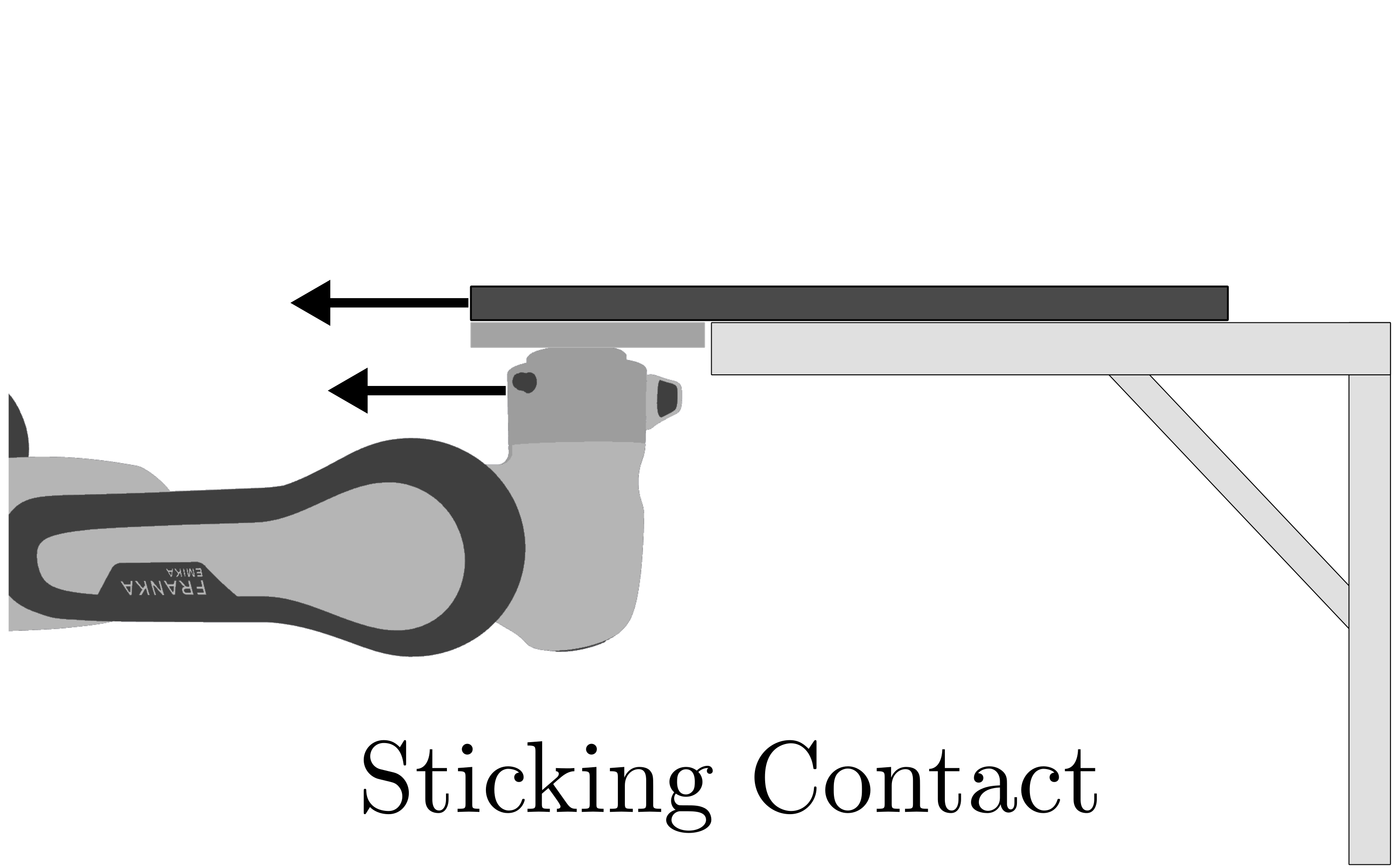}
	\includegraphics[height=6.7em]{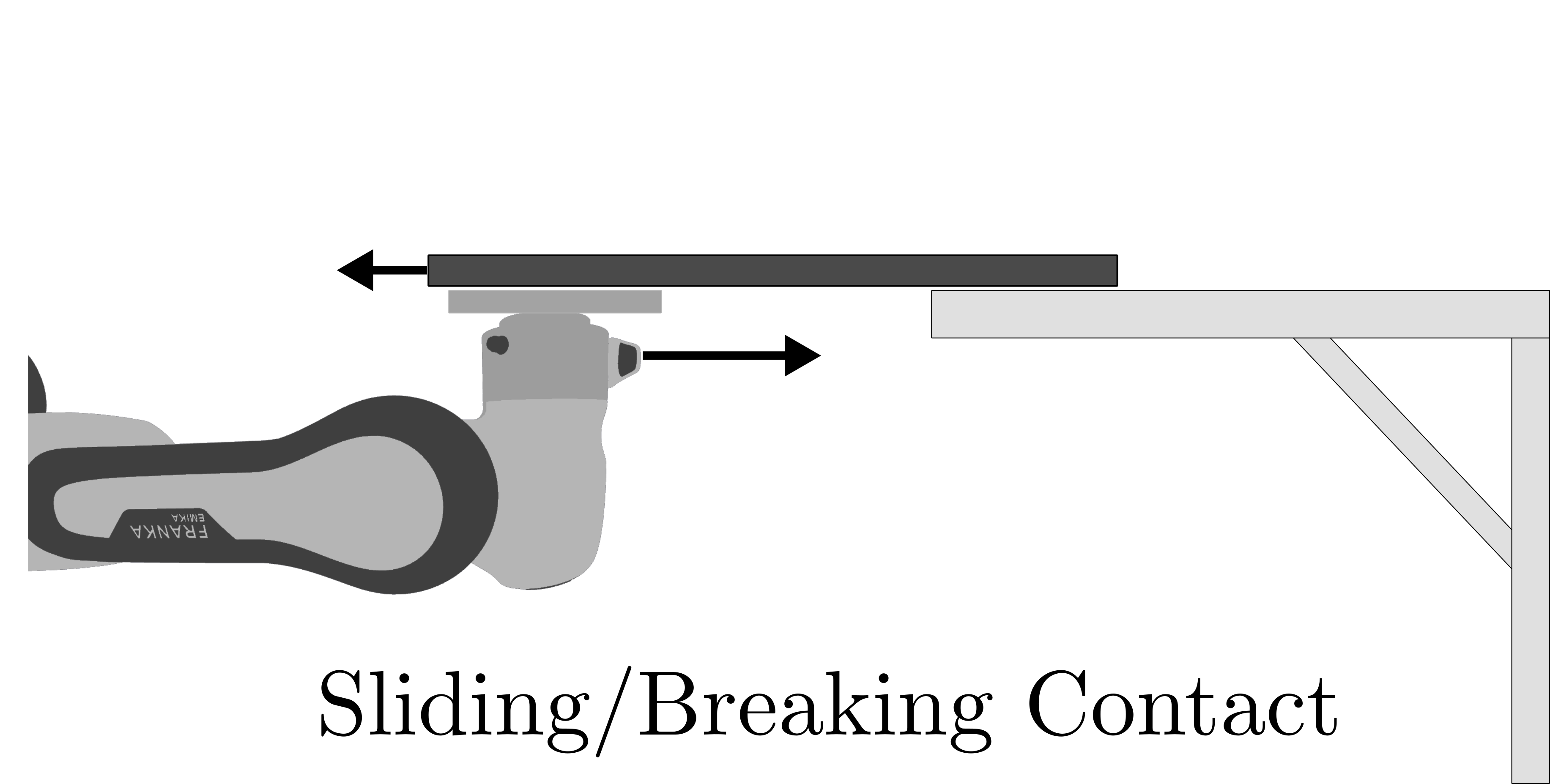}
	\caption{We examine a dynamic sliding task, where the robot uses the full spectrum of contact modes (sticking, sliding, making and breaking contact) in order to retrieve a tray resting on external supports. We use contact-implicit MPC to automatically plan when and where to use different contact modes. With careful consideration on how to integrate the simplified MPC model with the robot arm, we are able complete the entire maneuver of retrieving the tray, lifting it, and placing it back on the external supports in just 5 seconds, demonstrating dynamic capability for a contact-rich task.}
	\label{fig:task}
\end{figure}
\Revision{In this paper, we focus on an extension of the waiter's task to serve as an example of a general class of problems that involve multiple contacts, reasoning about external contact, as well as stick-slip behavior. This task both resembles real dexterous manipulation skills and also exemplifies a range of challenges faced in general-purpose dexterous manipulation.}
In contrast with prior renditions of the waiter's task \cite{subburaman2023non} \cite{heins2023keep} \cite{brei2023serving}, which focus on just transporting the tray while maintaining static contact, our task simulates the full process of first retrieving the tray, then lifting the tray, and finally placing the tray back at its initial position.
The tray initially rests on external supports, so that a portion of the tray hangs over the edges of the supports.
As illustrated in \cref{fig:task}, in order to retrieve the tray, the robot must first shift the tray so that it slides onto the end effector before it can be supported from underneath.
Similarly, in order to place the tray back at its initial position, the end effector must shift the tray forward onto the supports.
Both of these maneuvers require repeated stick-slip transitions.
The primary challenge of this task is the consideration of dynamic frictional contact.
One major challenge of frictional contact is the known model inaccuracies of Coulomb friction and rigid frictional impacts \cite{remy2017ambiguous}.
Another major challenge is that sliding adds additional contact modes to the already challenging hybrid planning problem.
Prior works that consider control with sliding contact are restricted to a single contact for planar dynamic tasks \cite{shi2017dynamic} \cite{hou2020robust} or multiple contacts for planar quasi-static tasks \cite{doshi2022manipulation} \cite{taylor2023object}.
\Revision{Other methods can reason about sliding contacts in 3D quasi-static and quasi-dynamic tasks \cite{cheng2022contact} \cite{cheng2023enhancing}; however, the methods are currently too slow for real-time control.}

Surprisingly, we show that, with some improvements, a general contact-implicit model predictive controller (MPC) framework can accomplish this dynamic task. Specifically, we build upon prior work by carefully considering the integration between the simple model used by the MPC and the low-level tracking controller in order to accurately track the dynamic motions commanded by the MPC.
Our controller automatically plans motions with repeated stick-slip transitions as the robot pushes or pulls the object, including initiating slipping to reposition the end effector to then push or pull again.
The controller accomplishes this all without using heuristics or commonly relied on aids such as reference trajectories or motion primitives.
This work makes the following contributions:
\begin{itemize}
	\item Proposal and demonstration of a complex object pick and place task that requires regulating sticking and sliding contact modes as well as making and breaking contacts.
	\item Improvements to contact-implicit MPC including integration with the downstream tracking controller to accurately track dynamic motions.
	\item Extensive experimental validation of the proposed framework and hypotheses for characteristics of robust stick-slip maneuvers.
	\Revision{\item Demonstration of the generality of the contact-implicit MPC framework with a task where the robot must rotate a circular tray using a wall.}
\end{itemize}

\section{Related Work}
\label{sec:related_works}

\subsection{Contact Mode Regulation}

The primary exploration of this work is how to plan and regulate between frictional contact modes (sliding, sticking, and breaking contact altogether), with an emphasis on dynamic sliding contact as it is comparatively unexplored.
Dynamic sliding \cite{shi2017dynamic} and pivoting \cite{hou2020robust} for object reorienting and regrasping are performed by simultaneously regulating inertial and frictional forces using a parallel jaw gripper. 
However, these works are limited to the planar case and only consider a single surface contact.
\citet{doshi2022manipulation} demonstrate impressive control of both sliding and sticking contact along multiple surfaces including utilizing external contacts \cite{taylor2023object}; however, they focus on quasi-static manipulation and again are limited to a planar system.
\citet{woodruff2017planning} demonstrate sequencing multiple motion primitives \cite{lynch1999dynamic}, including dynamic sliding, demonstrated on a planar manipulator and block set up. However, the full trajectory is planned offline and relies on local feedback control to stabilize each motion primitive.

We highlight the planar nature of prior examples because planning for sliding contact in 3D is fundamentally more challenging than in the planar case.
This is because, in addition to the increased state dimension, the planar case only requires consideration of 4 hybrid modes (sticking, no contact, slide left, slide right) per contact, whereas there exists a \textit{continuum} of sliding modes for the 3D case.
\citet{higashimori2009dynamic} considers full surface-surface friction when manipulating a flat object with 3 degrees of freedom (DOFs) resting on a pizza peel-like platform with only controlled 2 DOFs.
Their work showcases impressive controllability, but the method assumes that the platform is much larger than the object and the nominal pressure distribution on the object being uniform.

\subsection{Waiter Task}
Several works \cite{pham2017admissible} \cite{heins2023keep} \cite{subburaman2023non} \cite{brei2023serving} have tackled the ``waiter task", where objects are balanced on top of an end effector with a planar surface.
Despite the similar task set-up, all of these works critically focus on avoiding sliding between the object and manipulator, whereas the key focus of our work is to specifically leverage sliding to perform tasks that would otherwise be infeasible.

\subsection{Contact-Implicit MPC}

Recently, several contact-implicit MPC frameworks have demonstrated solve times fast enough for real-time control on systems with multiple contacts and many degrees of freedom \cite{aydinoglu2023consensus} \cite{kurtz2023inverse} \cite{le2024fast}.
However, these methods have not been evaluated on dynamic manipulation with sliding contacts.
Here, we do not propose a new MPC framework, but rather we seek to identify the implementation details to adapt such a framework to a dynamic task, including how to consume the outputs in a downstream tracking controller.
Critically, the output of these contact-implicit methods were tracked as position set-points stabilized with impedance gains \cite{aydinoglu2023consensus} \cite{kurtz2023inverse}, whereas we track time-parameterized trajectories for the end effector position and end effector forces.
Position set-points rely on the stiffness of the impedance controller to achieve accelerations, while accelerations and forces can be specified directly and are defined smoothly using time-parameterized trajectories.

\section{Problem Setup}
\label{sec:problem_setup}

\begin{figure}[h]
	\centering
	\includegraphics[width=0.48\textwidth]{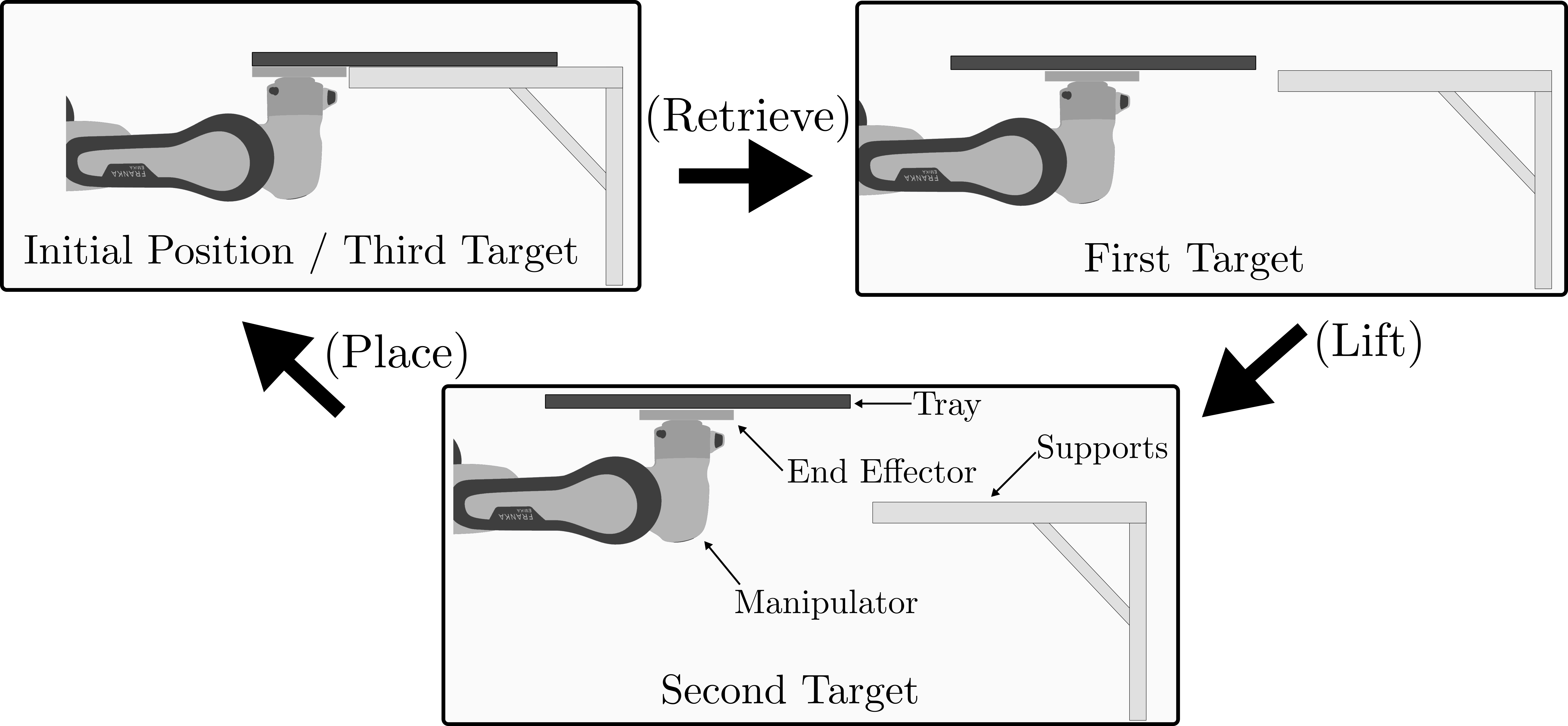}
	\caption{The three target positions. The grasp locations on the tray change between targets, thus requiring the end effector to either slide and/or break contact with the tray.}
	\label{fig:target_positions}
\end{figure}

We are interested in the problem setup shown in \cref{fig:target_positions}.
The system consists of a serial-link manipulator equipped with a small flat disk as its end effector, where the end effector is constrained to move only in 3D translation.
The robot arm is tasked with retrieving, lifting, and returning (placing) a tray as shown in \cref{fig:target_positions}.
The tray is initially resting on external supports and starts in a slightly overhanging position so that it can be contacted on its bottom surface.
The tray has all floating base degrees of freedom, with its pose in $SE(3)$, and its pose can be tracked using fiducials attached to the tray.
We assume that we have accurate model parameters (mass, inertia, geometry, and friction) of each component (robot arm, end effector, tray, and supports), although we do examine the effect of inaccurate models in \cref{sec:results}.
We assume a single coefficient of friction per pair of geometries (tray/end effector and tray/supports).
The task objectives: retrieve, lift, and place, are specified as three sequential targets, meaning the next target is given when the tray reaches the previous target.

\section{System Models}

\begin{figure}[h]
	\includegraphics[width=0.48\textwidth]{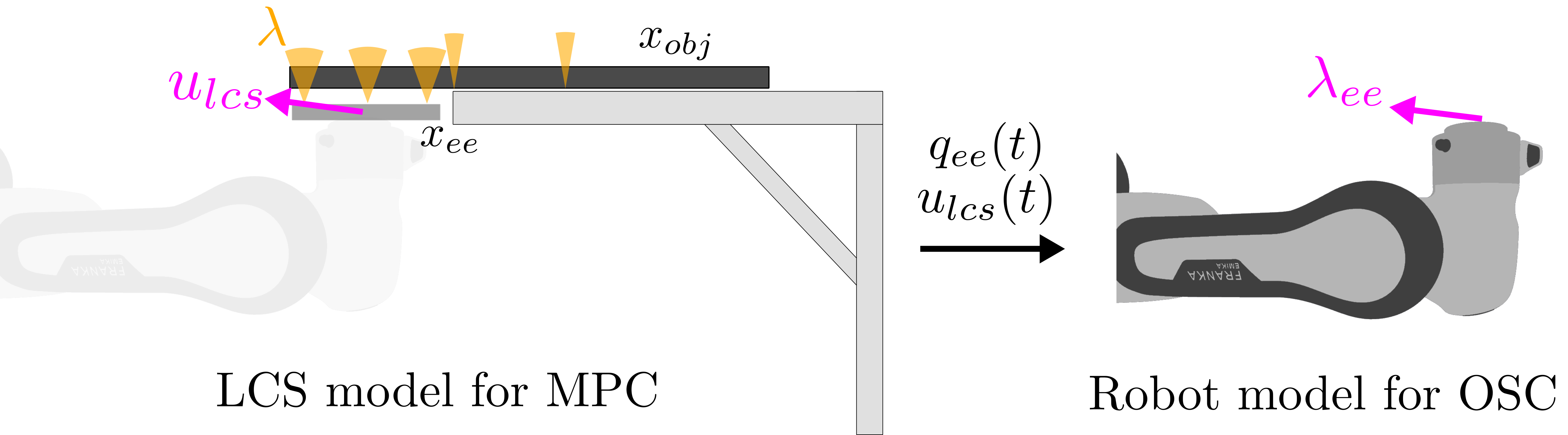}
	\caption{We abstract the system into two models. The LCS model captures the contact forces $\lambda$ between the end effector, tray, and supports. In the LCS model, the robot arm is abstracted away and replaced with direct inputs to the end effector $u_{lcs}$. We then use a robot-only model to track the end effector position $q_{ee}(t)$ and force $u_{lcs}(t)$ trajectories commanded from the MPC, so $\lambda_{ee} = u_{lcs}$.}
	\label{fig:task_models}
\end{figure}

In this paper, we abstract the system using two models as shown in \cref{fig:task_models}.
We model the the end effector, object, and external contacts as a Linear Complementarity System \cite{heemels2000linear} (LCS) to use for the MPC.
For the low-level operational space controller, we only consider the dynamics of the robot arm and rely on inputs from the MPC to address the interaction forces from the object.

\subsection{Linear Complementarity Model}

We model the dynamics of the end effector, object, and external contacts as a discrete time LCS.
We ignore rest of the robot arm in the MPC model by considering the end effector as an isolated floating object with only translation degrees of freedom and controlled directly with forces applied to its center of mass.
To ensure downstream feasibility when applying this model to the actual system, we impose workspace and input limits on the MPC model.
The state of the LCS $x_{lcs} = [q_{lcs}, v_{lcs}]$ is a combination of the positions $q_{lcs} = [q_{ee}, q_{obj}]$ of the end effector and object and the corresponding velocities $v_{lcs} = [v_{ee}, v_{obj}]$.
The control input $u_{lcs}$ to the LCS are forces applied directly to the end effector center of mass, such that it can be controlled in 3D translation.
Finally, the contact forces $\lambda$ are the interaction forces between the end effector, object, and external supports.

The dynamics of the LCS have the form:
\begin{align}
	x_{k+1} = A x_k + B u_k + D \lambda_k + d \label{eq:lcs_dynamics}\\
	0 \leq \lambda_k \perp E x_k + F \lambda_k + H u_k + c \geq 0 \label{eq:lcs_contact},
\end{align}
where $x_k \in \Real^{n_{x}}$, $\lambda_k \in \Real^{n_{\lambda}}$, and $u_k \in \Real^{n_u}$ are the state, force, and input variables at the $k$-th knot point.
\Revision{\cref{eq:lcs_dynamics} are the system dynamics linearized at the current state and input.}
The $\perp$ indicates a complementarity constraint, where $0 \leq \lambda \perp \phi \geq 0$ implies $\lambda \geq 0$, $\phi \geq 0$, $\lambda^T \phi = 0$.
Critically, this constraint succinctly describes the multi-modality of contact dynamics for both making and breaking contact as well as the boundary between stick and slip.
For example, the boundary between sliding and sticking friction for a given sliding direction can be expressed as:
\begin{align}
	0 \leq \mu \lambda_n - \lambda_t \perp v \geq 0,
\end{align}
where $\lambda_n$ is the normal force, and $\lambda_t$ is the tangential force in the opposite direction of the sliding velocity $v$.
\Revision{With this context, \cref{eq:lcs_contact} is the linearization of the contact boundaries at the current state and input.}

\begin{figure}
	\includegraphics[width=0.48\textwidth]{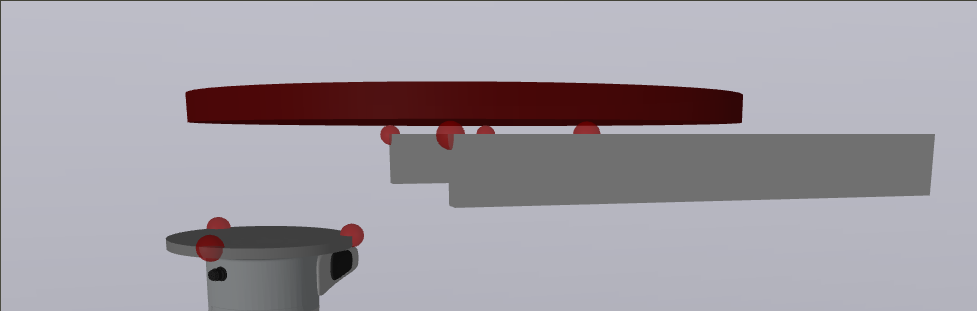}
	\caption{We consider seven total contacts for our task. The contact geometries shown in red. We represent the tray as a cylinder and we choose fixed contact points on the end effector and supports, which we model as spheres. The radii for the spheres are enlarged by a factor of 10 for visibility purposes. A minimum of three contact points are required to approximate surface-surface contact between the end effector and tray, while two contact points are required to model each line contact from the supports.}
	\label{fig:contact_geometries}
\end{figure}

The contact dynamics of the LCS are governed by our choice of contact geometry.
We approximate the surface-surface contacts between the end effector and object as well as the object and external contacts using point contacts.
We use three contact points between the end effector and object, because that is the minimum number necessary to have statically stable surface-surface contact.
Similarly, we model each support as two points to represent the line contacts.
These contact geometries are visualized in \cref{fig:contact_geometries}.
Under this choice of contact geometry, $\phi$ encodes the distance between any of these contact points and the tray, represented as a cylinder.
Under this modeling choice, the number of contacts $n_{\lambda}$ is fixed.
Note, we are ignoring potential contacts between the end effector and the supports.
As these contacts are undesirable, we simply avoid these contacts by imposing workspace constraints on the end effector.

\subsection{Robot-Only Model}
\label{subsec:osc_dynamics}

We only consider the state of the arm when applying our low level tracking controller.
We denote the state of the robot arm as $x_{arm} = [q_{arm}, \dot q_{arm}]$, which is comprised of its generalized positions $q_{arm} \in \Real^{n_{arm}}$ and generalized velocities $\dot q_{arm} \in \Real^{n_{arm}}$.
The arm is controlled using actuator inputs $u_{arm} \in \Real^{n_{arm}}$, where the inputs are motor torque commands.
For brevity, we omit the $_{arm}$ subscript for the remainder of this section.
We can describe the arm dynamics using the manipulator equation:
\begin{align}
	M(q) \ddot{q} + C(q, \dot q) = B u + J(q)^T \lambda_{ee},
	\label{eq:robot_dyn}
\end{align}
where $M$ is the mass matrix with approximated reflected inertia terms \cite{featherstone2014rigid}, $C$ contains the Coriolis and gravitational forces, $B$ maps actuator inputs to generalized forces, and $J$ is the contact Jacobian that maps forces $\lambda_{ee}$ applied at the end effector to generalized forces.

\section{Methods}
\label{sec:methods}

\subsection{Complementary Consensus Control}
\label{subsec:mpc}

We formulate our control problem as a contact-implicit MPC problem with LCS dynamics.
This is succinctly formulated as the following optimization problem:
\begin{align}
	\min_{x_k, u_k, \lambda_k}&& x_N^T Q_f x_N + \sum_{k=0}^{N-1} x_k^T Q x_k + u_k^T R u_k \\
	s.t. && 	x_{k+1} = Ax_k + D\lambda_k + B u_k + d \label{eq:lcs_dynamics_constraint}\\
	&& E x_k + F \lambda_k + H u_k + c \geq 0 \label{eq:linearized_gap_function}\\ 
	&& \lambda_k \geq 0 \label{eq:positive_force_constraint}\\
	&& \lambda_k^T (E x_k + F \lambda_k + H u_k + c) = 0 \label{eq:orthogonal_constraint}\\
	&& x_{min} \leq x_k \leq x_{max}\\
	&& u_{min} \leq u_k \leq u_{max},
	\label{eq:mpc_lcs}
\end{align}
where $N$ is the planning horizon, $Q_f, Q, R$ are cost matrices, and $x_{min}, x_{max}, u_{min}, u_{max}$ are bounds on the state and input variables.
\sout{The non-convex constraint \mbox{\cref{eq:orthogonal_constraint}} can be converted to binary variables and the entire problem can be solved as a Mixed Integer Quadratic Program (MIQP).}
\Revision{\cref{eq:lcs_dynamics_constraint}, \cref{eq:linearized_gap_function}, and \cref{eq:positive_force_constraint} are the dynamics constraints of the LCS. \cref{eq:orthogonal_constraint} is the orthogonality constraint for the complementarity.
Note, \cref{eq:orthogonal_constraint} is non-convex, but it is possible to introduce binary variables to represent the contact modes and transcribe the entire problem as a Mixed Integer Quadratic Program (MIQP).
However, this scales poorly with the number of contacts, as a binary variable is needed for each contact across all knot points.}

Instead, we adopt a method called Complementarity Consensus Control (C3) \cite{aydinoglu2023consensus}, which addresses the scaling problem by decoupling the time dependence of the contact decisions.
The algorithm is based in consensus alternating direction method of multipliers (ADMM), which optimizes over multiple copies of the decision variables.
The full details of the algorithm is outside of the scope of this paper, but a key property is that the algorithm alternates between solving the MPC problem as a quadratic program (QP) without complementarity constraints and projecting the current MPC solution to the complementarity constraints as a mixed integer quadratic program (MIQP) \textit{separately} for each knot point.
While the solutions will eventually converge to each other, we choose to terminate early after a fixed number of iterations on a potentially suboptimal solution.
We choose to terminate after the QP step, because we empirically observe better performance.
Note, the suboptimal solutions from terminating early do not necessarily satisfy the full LCS dynamics, but in practice are good enough when used in MPC for even contact-rich tasks.

\subsection{System Linearization}

\begin{figure*}[ht]
	\includegraphics[width=0.33\textwidth]{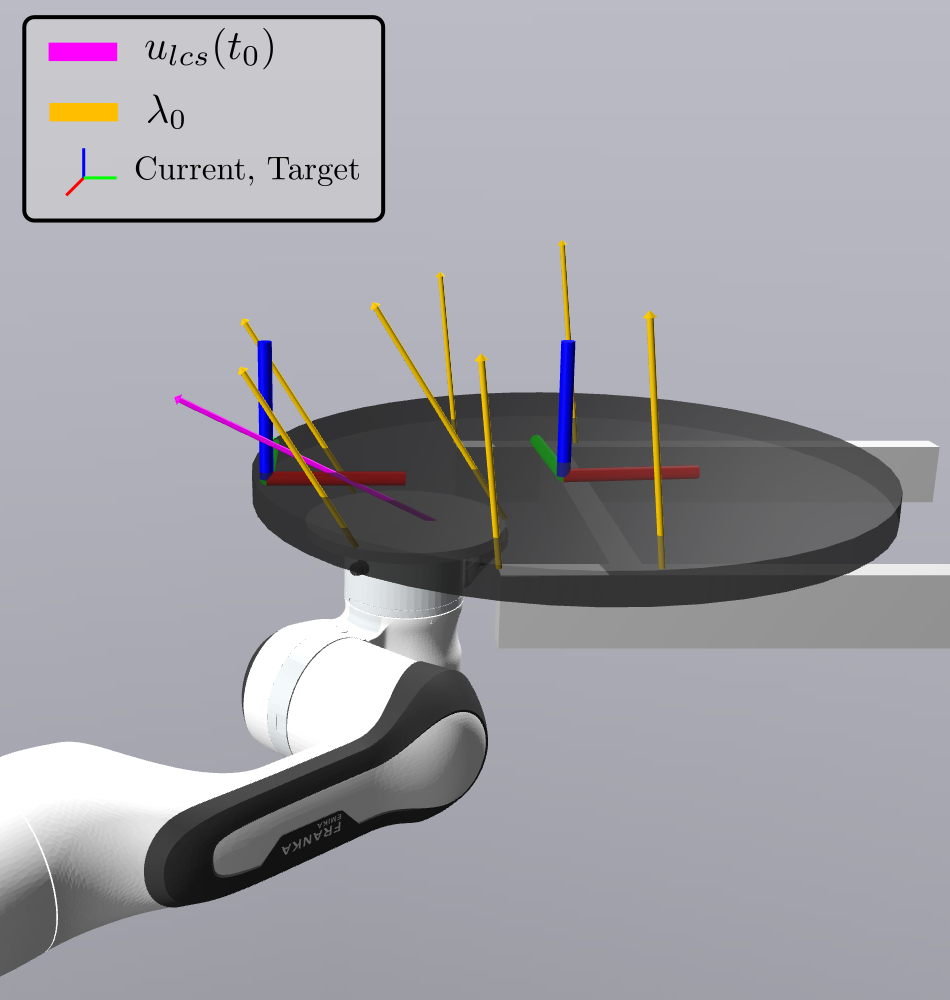}
	\includegraphics[width=0.33\textwidth]{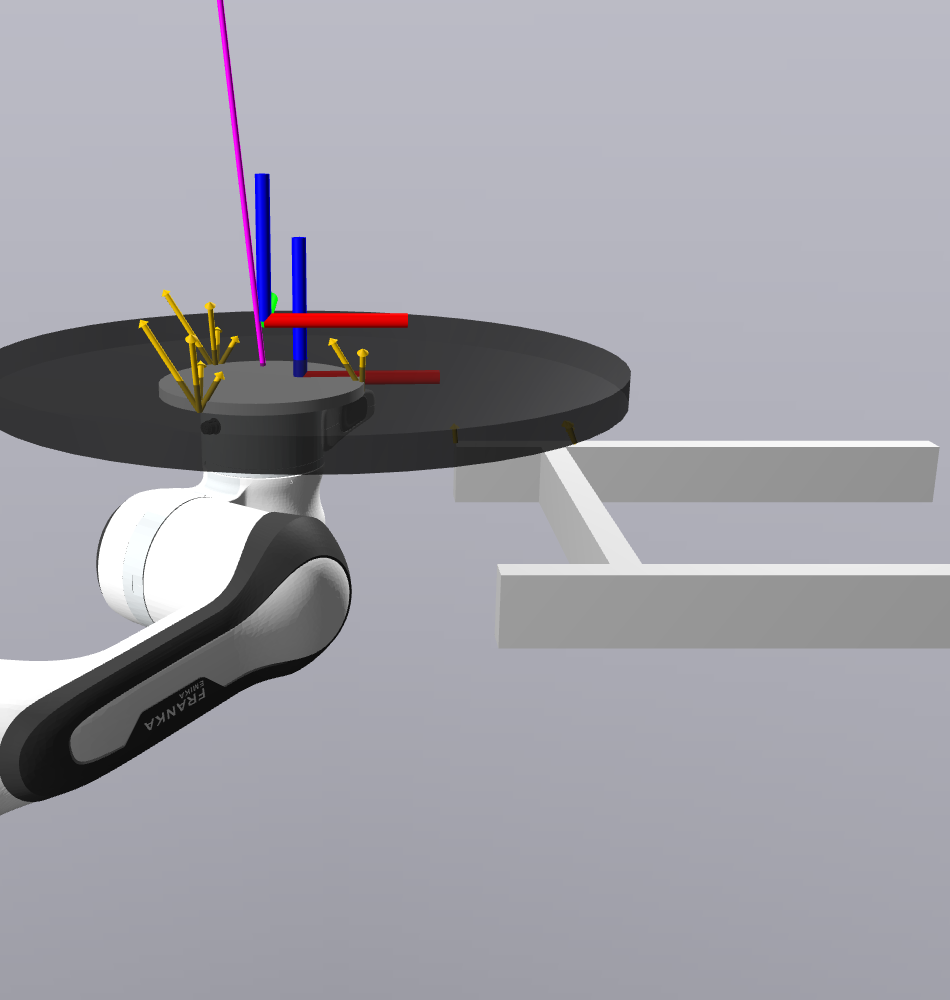}
	\includegraphics[width=0.33\textwidth]{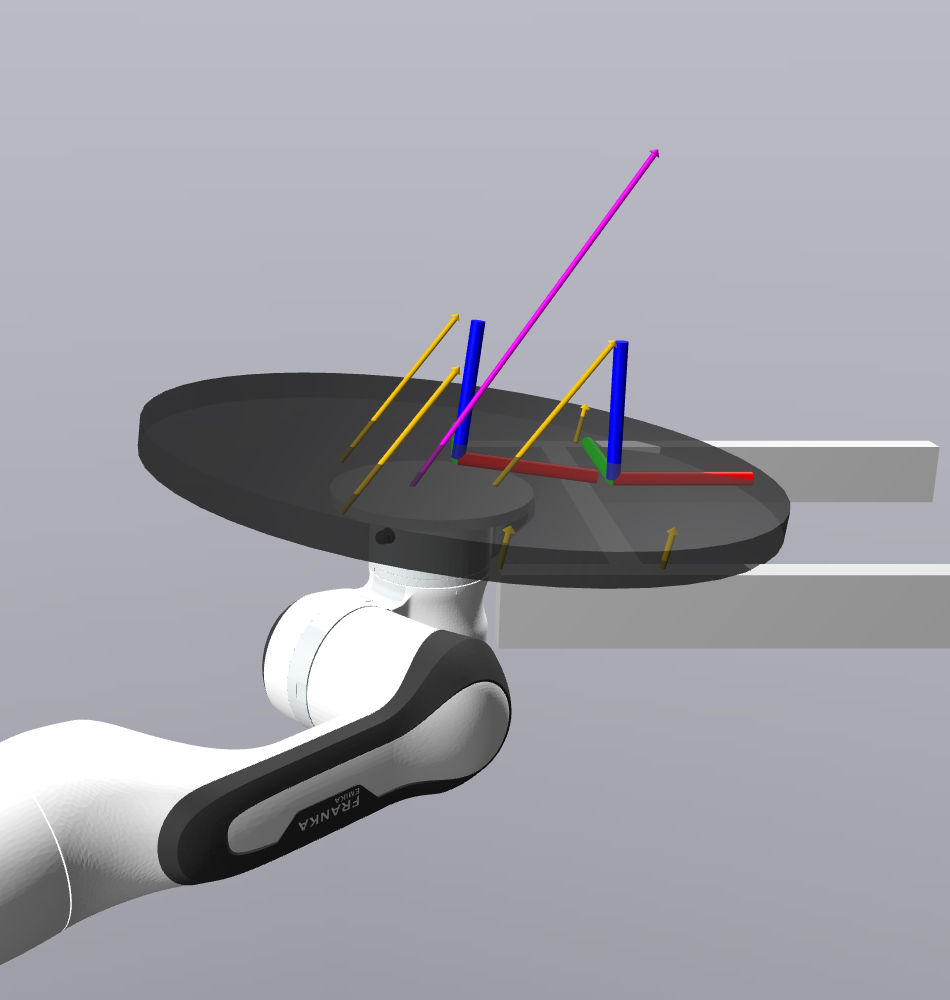}
	\includegraphics[width=0.33\textwidth]{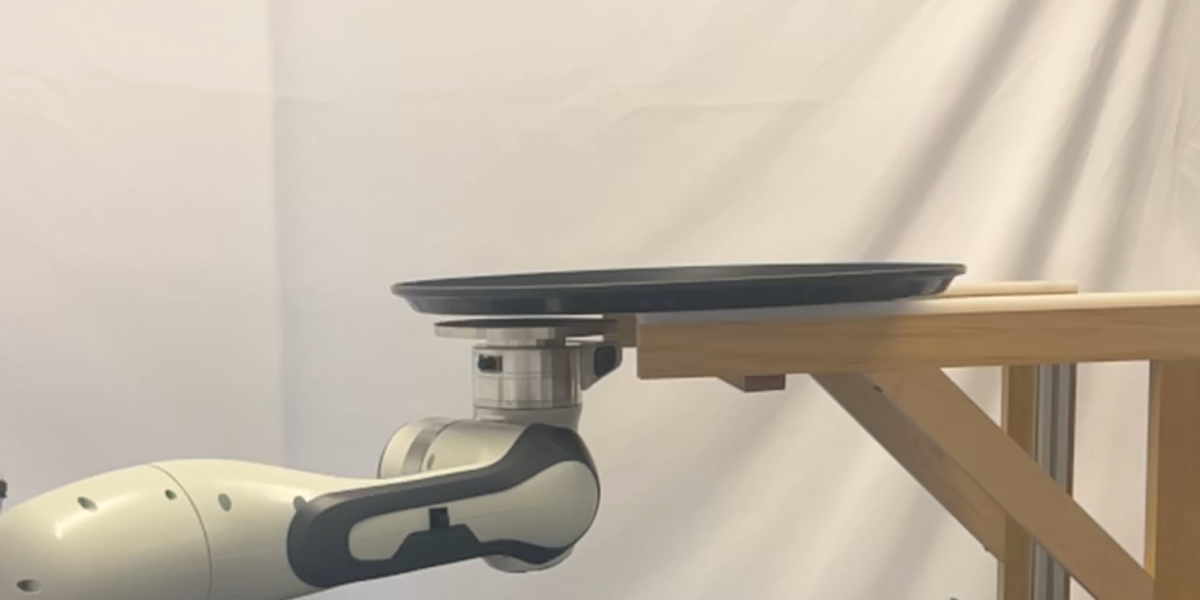}
	\includegraphics[width=0.33\textwidth]{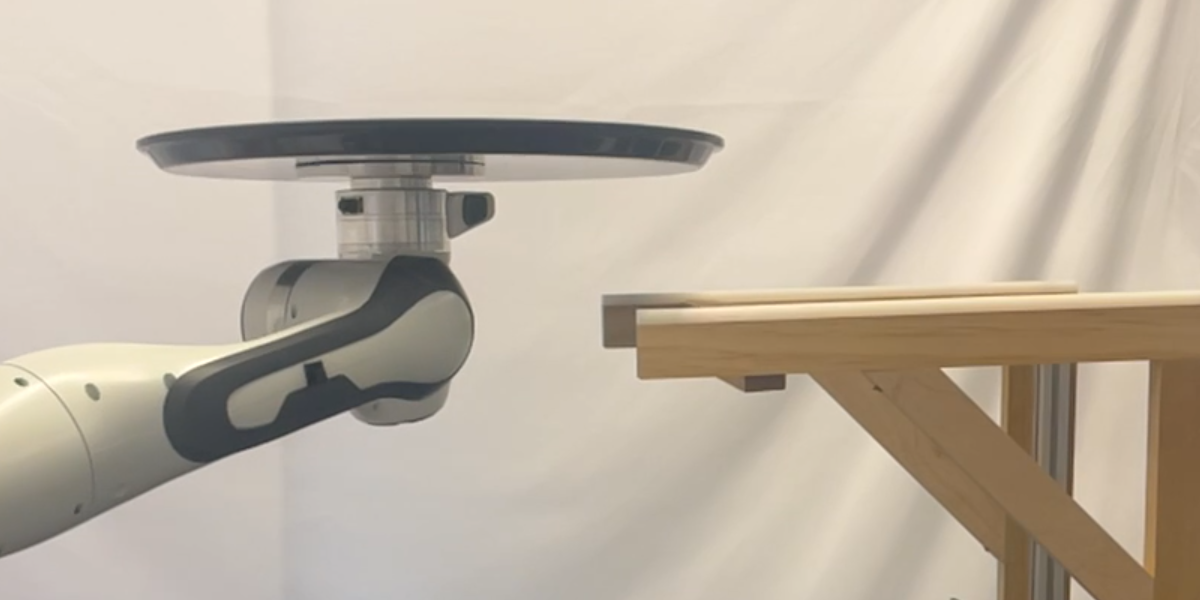}
	\includegraphics[width=0.33\textwidth]{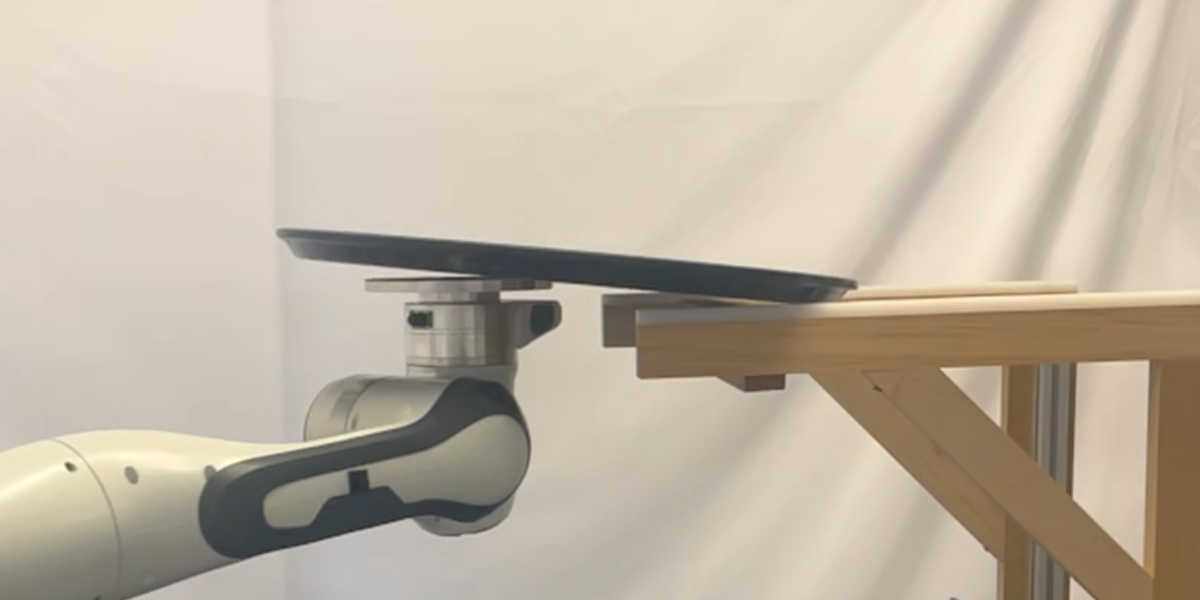}
	
	\caption{Examples of the MPC plan for retrieval (left), lift (center), and place (right), where the current state and target state of the tray are represented as triads. The MPC plans the states, inputs, and forces for $N$ timesteps into the future. The forces (yellow arrows) and inputs (pink arrow) at the first timestep are visualized for each maneuver. For the retrieval maneuver, the plan heavily relies on the external supports compared to the other two maneuvers, where the primary contacts forces are between the end effector and tray.}
	\label{fig:c3_plan}
\end{figure*}

We approximate our system as a LCS at each C3 solve.
The continuous dynamics parameters ($A$, $B$, $d$) of the LCS can be solved via automatic differentiation using any popular rigid body dynamics library.
The gap function $\phi$ and corresponding contact Jacobians $J$ for convex geometries can be computed by a library that implements collision detection, e.g. via the GJK algorithm\cite{gilbert1988fast}.
With $\phi$ and $J$ for each contact along with a choice of force basis, we can compute the contact-related terms: $D$, $E$, $F$, $H$, $c$.
We use the convex time-stepping contact model proposed by \citet{anitescu1997formulating} to form the force basis.
In this model, contact forces are parameterized via the extreme rays of the pyramidal approximation of the friction cone.
That is, for a square pyramidal approximation, there are 4 contact force variables per point contact.
This choice of contact force basis is visualized in \cref{fig:c3_plan}.

\subsection{MPC Modifications for Dynamic Motions}

The fast motions commanded by our MPC are on the same timescale as the solve time.
For this reason, the system state at the end of the MPC solve is likely far from the system state at the beginning of the MPC solve.
We address this latency problem by using the predicted state of the system according to the previous MPC solve as the initial state constraint similar to \cite{bledt2020regularized}
\begin{align}
	x_0 &= x_{sol}(\overline{dt}),
\end{align}
where $x_{sol}(t)$ is the state trajectory from the previous MPC solve and $\overline{dt}$ is the filtered average MPC solve time.
\Revision{Because we have less confidence in the accuracy of our contact models, we only apply this prediction to the end effector state and not the state of the tray.}

Warm starting by giving the MPC an initial guess from the previous solve is a common technique to reduce computation time.
\Revision{However, the values from the previous solution are often poor initial guesses because the contact modes at each knot point planned from the current MPC state may differ from the previous solution. Because the dynamics can vary greatly between contact modes, so can the values for $x, u, \lambda$. For this reason, we use the corresponding predicted values from the previous solution when possible for warm starting.}
Additionally, C3 involves multiple QP and MIQP solves per MPC solve each with different cost parameters.
We address this by treating each QP and MIQP as separate optimization programs each with a separately cached set of warm start variables in order to increase the quality of the warm start.

\subsection{Operational Space Control}
\label{subsec:low_level_control}

We use a low-level tracking controller to stabilize the plans commanded by the MPC. 
Specifically we track the end effector position, orientation, and force applied at the end effector specified as time-parameterized trajectories.
To achieve these accelerations in our low-level tracking controller, we use an operational space controller (OSC) \cite{khatib1987unified} \cite{wensing2013generation}, which is an inverse dynamics controller designed to track task-space objectives. 

It accomplishes this by finding the optimal actuator torques that best tracks the commanded task space accelerations $\ddot y_{cmd, i}$, which is computed as the desired end effector accelerations $\ddot y_{des, i}$ stabilized with PD gains in task space formulated as
\begin{align}
\ddot y_{cmd, i} = \ddot y_{des, i}(t) + K_p (y_{des, i}(t) - \taskSpacePos) + K_d (\dot y_{des, i}(t) - \taskSpaceVel).
\label{eq:task_space_acc_cmd}
\end{align}
We directly track the end effector force objective \sout{$\lambda_{des}$} \Revision{$\lambda_{ee}$}.

We formulate this as the following quadratic program (QP):
\begin{align}
	\min_{u, \lambda, \ddot q}   &\quad&  \Norm{\lambda - \Revision{\lambda_{ee}}}^2_W  + \sum_{i}^{N} \Norm{\taskSpaceAccErr_i}^2_{W_i} && & \label{eq:osc_qp_full}\\
	\text{s.t. }
	&\quad&  M \genAcc + C = u + \contactJacobian^T \lambda  &&  & \label{eq:dyn_constraint_full} \\ 
	&\quad& u_{min} \leq u \leq u_{max}, &&  \label{eq:actuator_limits_upper}
\end{align}
where $i$ refers to each task space objective and whose acceleration $\ddot y_i$ is linear mapping of $\ddot q$ and can be derived from the task space kinematics function $\psi_i$, where $y = \psi_i(q)$. 
Differentiating with respect to time, we have $\dot y = \frac{\partial \psi}{\partial q} \dot q = J_{y, i} \dot q$, and differentiating once more we get $\ddot y = \dot J_{y, i} \dot q + J_{y, i} \ddot q$.
\cref{eq:dyn_constraint_full} is the dynamics constraint that relates actuator inputs $u$ to joint accelerations $\ddot q$ and the actuator lower and upper limits are specified by $u_{min}$ and $u_{max}$ respectively.

Using this general OSC formulation, we explicitly define the tracking objectives as follows.
The time-varying objectives of the OSC are the end effector position trajectory $q_{ee}(t)$ and the end effector force trajectory $u_{lcs}(t)$, so $y_{des, 0}(t) = q_{ee}(t)$ and corresponding derivatives and $\Revision{\lambda_{ee}} = u_{lcs}(t)$.
For the other objectives, the end effector orientation target is the neutral quaternion $y_{des, 1} = [1, 0, 0, 0], \dot y_{des, 1} = \ddot y_{des, 1} = [0, 0, 0]$ because we assume the end effector can only move in translation degrees of freedom.
Additionally, in order to keep the robot arm in the ``elbow down" configuration and have a unique robot configuration for a given end effector position and orientation target, we specify a single joint-space tracking objective to keep the second joint of the arm at a fixed angle.

\subsubsection{End Effector Force Target}

The end effector force target is an important component to accurately tracking the MPC plan without relying on overly stiff impedance gains or an integral term, both of which could cause instability for this task.
To see this, consider the scenario where the robot balances the tray. 
Without a force target, the robot will not compensate for the weight of the object, and the object will sag according to the impedance stiffness $K_p$.
While tracking error for interactions solely between the manipulator and object scales with stiffness, tracking error for systems with additional contacts is more complex.
For example during the sliding maneuver, even small forces applied by the end effector to the object can result in significant effects on the weight distribution of the object across the supports and end effector.
Because our task is governed by friction forces, this objective is particularly important.

\section{Experiments}
\label{sec:experiments}

\begin{figure*}[ht!]
	\centering
	\includegraphics[width=0.98\textwidth]{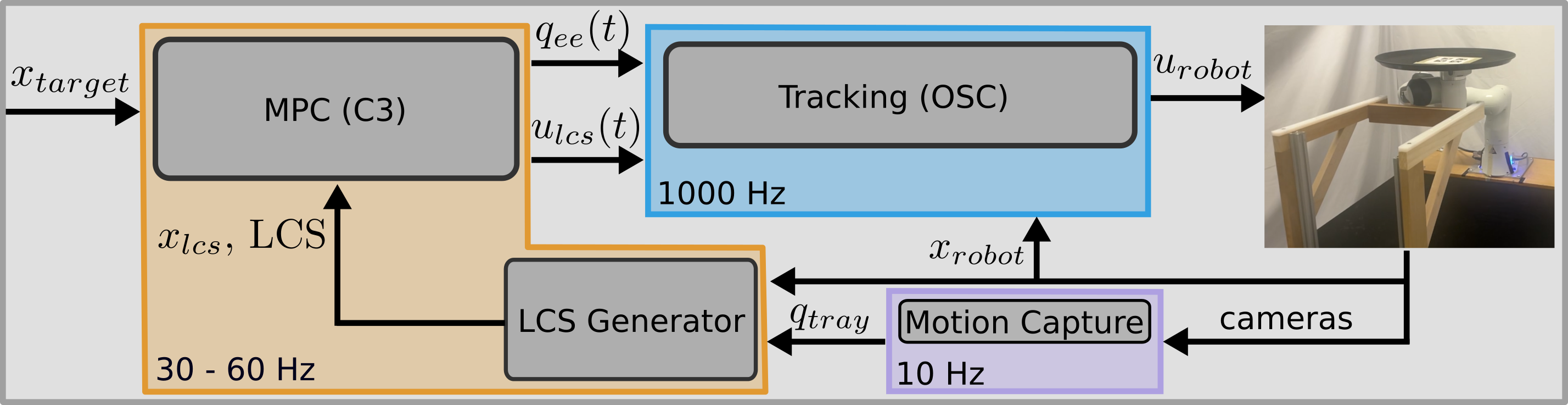}
	\caption{System diagram for the hardware implementation. The different colored boxes indicate separate processes which are connected via arrows that indicate represent communication via ROS/LCM.}
	\label{fig:system_diagram}
\end{figure*}

\subsection{Task Parameters}

The exact position targets are given in \cref{tab:target_positions}. The positions are defined in the world frame where the robot base is at the origin.
For the first and second targets, the end effector target is the same as the tray, just offset in the vertical position to compensate for the thicknesses.
Critically, the third target for the tray is outside the workspace limits of the end effector, so for this pair of targets the end effector target is chosen as the closest point within the feasible workspace.
We choose to make the third target to be identical to the initial position in order to be able to repeatedly execute the experiment without manually resetting the task environment.
As detailed in \cref{sec:problem_setup}, the next target is only given when the tray reaches the previous target.
We define reaching the target as being within 5 cm from the target location.

\begin{table}[h]
	\centering
	\begin{tabular}{|c|c|c|c|}
		\hline
		& Tray (m) & End Effector (m) & Idle Time (s)\\
		\hline
		Initial Position & [0.7, 0.0, 0.485] & [0.55, 0, 0.45] & \\
		First Target & [0.45, 0, 0.485] & [0.45, 0, 0.47 & 0.5 \\
		Second Target & [0.45, 0, 0.60] &  [0.45, 0, 0.585] & 3.0\\
		Third Target & [0.7, 0, 0.485] & [0.6, 0, 0.47] & \\
		\hline
	\end{tabular}
	\caption{Target positions for tray retrieval task. Positions are specified as meters and in the robot/world frame where the base of the robot is at the origin [0, 0, 0]. Idle time indicates how long the robot must remain at the target before the next target is given.}
	\label{tab:target_positions}
\end{table}

\subsubsection{Tray and End Effector}

\begin{figure}[h]
	\includegraphics[width=0.24\textwidth]{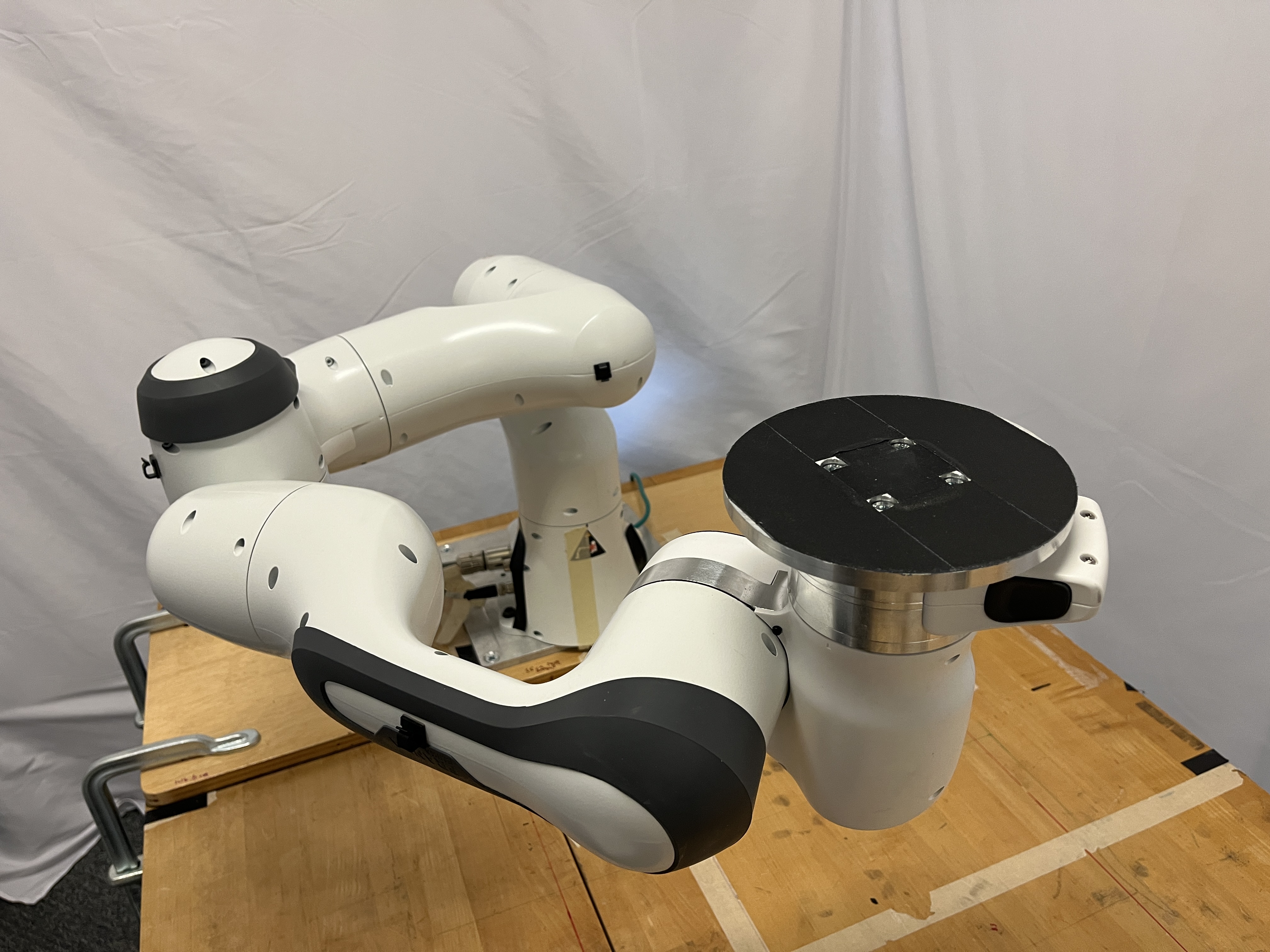}
	\includegraphics[width=0.24\textwidth]{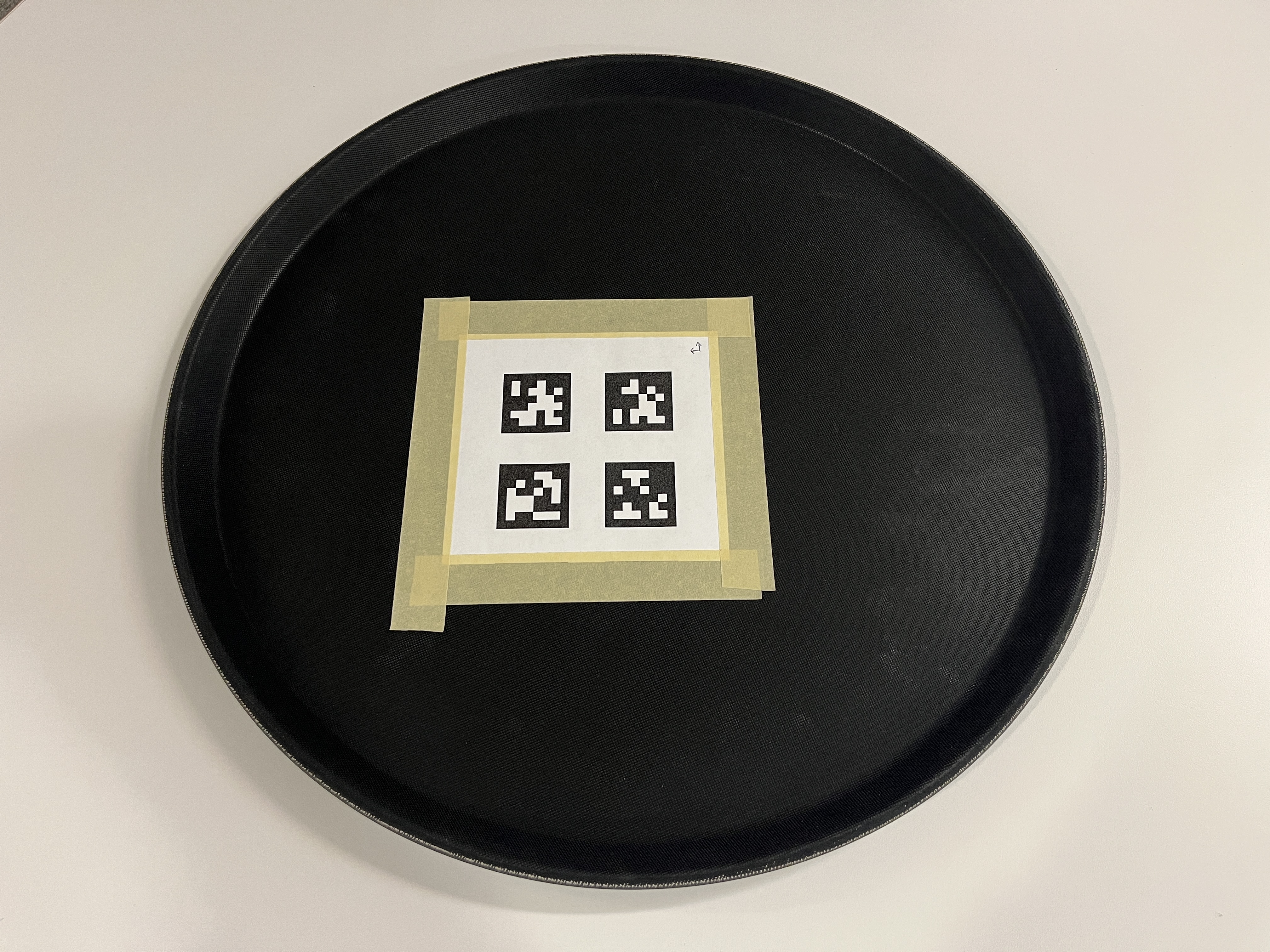}
	\caption{End effector attached to Franka robot and serving tray with attached AprilTag.}
	\label{fig:franka_hardware}
\end{figure}

We use a standard circular food service tray with a smooth low friction bottom surface and a rubberized high friction upper surface.
We model the tray as a cylinder with uniform density.
We machine the disk-shaped end effector out of aluminum.
Because the coefficient of friction between the machined aluminum and the tray's bottom surface is not sufficiently high, we cover the top surface of the end effector with tape.
We estimate the friction coefficients by slowly tilting the supports or end effector until the tray slips and using that angle to determine a single friction coefficient, assuming that the static and dynamic coefficients are identical.
Detailed parameters for both objects are listed in \cref{tab:physical_parameters} and the objects are shown in \cref{fig:franka_hardware}.

\begin{table}[h]
	\centering
	\begin{tabular}{|c|c|}
		\hline
		& Value \\
		\hline
		Tray Mass & 1 kg \\
		Tray Radius & 0.228 m \\
		Tray Thickness & 0.004 m \\
		Tray Height (including raised rim) & 0.022 m \\
		End Effector Mass & 0.37 kg \\
		End Effector Radius & 0.0725 m \\
		End Effector Thickness & 0.01 m \\
		Tray/Support Friction Coefficient & 0.18 \\
		Tray/End Effector Friction Coefficient & 0.5 \\
		\hline
	\end{tabular}
	\caption{Physical Parameters}
	\label{tab:physical_parameters}
\end{table}

\subsubsection{Franka Panda}

All communication between the simulator, C3 controller, and OSC were handled via LCM \cite{huang2010lcm}.
Communication between the low-level OSC controller and the Franka Panda was handled by a direct torque passthrough controller written using franka ros, a ROS wrapper around libfranka. 
We receive joint state messages from and send joint torques commands to the robot at 1000 Hz.
A separate LCM and ROS bridge is dedicated to translating between message types.
Notably, in franka ros, it was necessary to relax the torque and force thresholds from their default limits in order to accommodate the fast motions and interaction forces in this task.

\subsection{Implementation}

Both C3 and the OSC were implemented in C++ with the help of the Drake robotics library \cite{drake}.
Both controllers, as well as franka ros and the LCM to ROS bridges, are run on the same desktop with a Intel i7-8700K processor.
The QP step of C3 was solved using OSQP \cite{stellato2020osqp}, while the MIQP projection was solved using Gurobi \cite{gurobi}.
The OSC QP was solved using OSQP \cite{stellato2020osqp} at 1000 Hz.
We tune the OSC and C3 parameters by executing the task in the Drake \cite{drake} simulator using the hydroelastic contact model \cite{masterjohn2022velocity}.
We directly apply the parameters that were tuned in simulation on hardware without additional tuning.

Here, we report the most relevant parameters for C3 and leave the remaining parameters to be discussed in \cref{subsec:full_c3_parameters}.
We chose $N = 5$ knot points, a timestep of 0.075s for a time horizon of 0.3s, and 2 ADMM iterations for each C3 solve.
Under this choice of C3 parameters, we receive a new plan between 30 - 60 Hz.
The friction coefficient for the contacts between the tray and end effector was set to $\mu_{tray, ee} = 0.6$ and the friction coefficient for the contacts between the tray and the supports was set to $\mu_{tray, supports} = 0.1$.

\subsubsection{Motion Capture}
We use an off-the-shelf motion capture system \cite{pfrommer2019tagslam}, which uses AprilTags attached to the tray to publish the position of the tray via ROS at 10 Hz.

\section{Results}
\label{sec:results}

We performed multiple experiments to validate the robustness and generality of our framework.
First we ablate our design decision to include the force tracking objective in the OSC by running experiments with and without that objective.
We then demonstrate the reliability by continuously executing the experiment without manual resetting.
Then, we demonstrate the robustness of our method to inaccurate models of mass and inertia by placing objects on the tray.
We use the same controller parameters for all three targets and across all the demonstrations, and footage of the experiments can be seen in the supplemental video.

\subsection{Force Tracking Ablation}
\begin{figure}
	\includegraphics[width=0.48\textwidth]{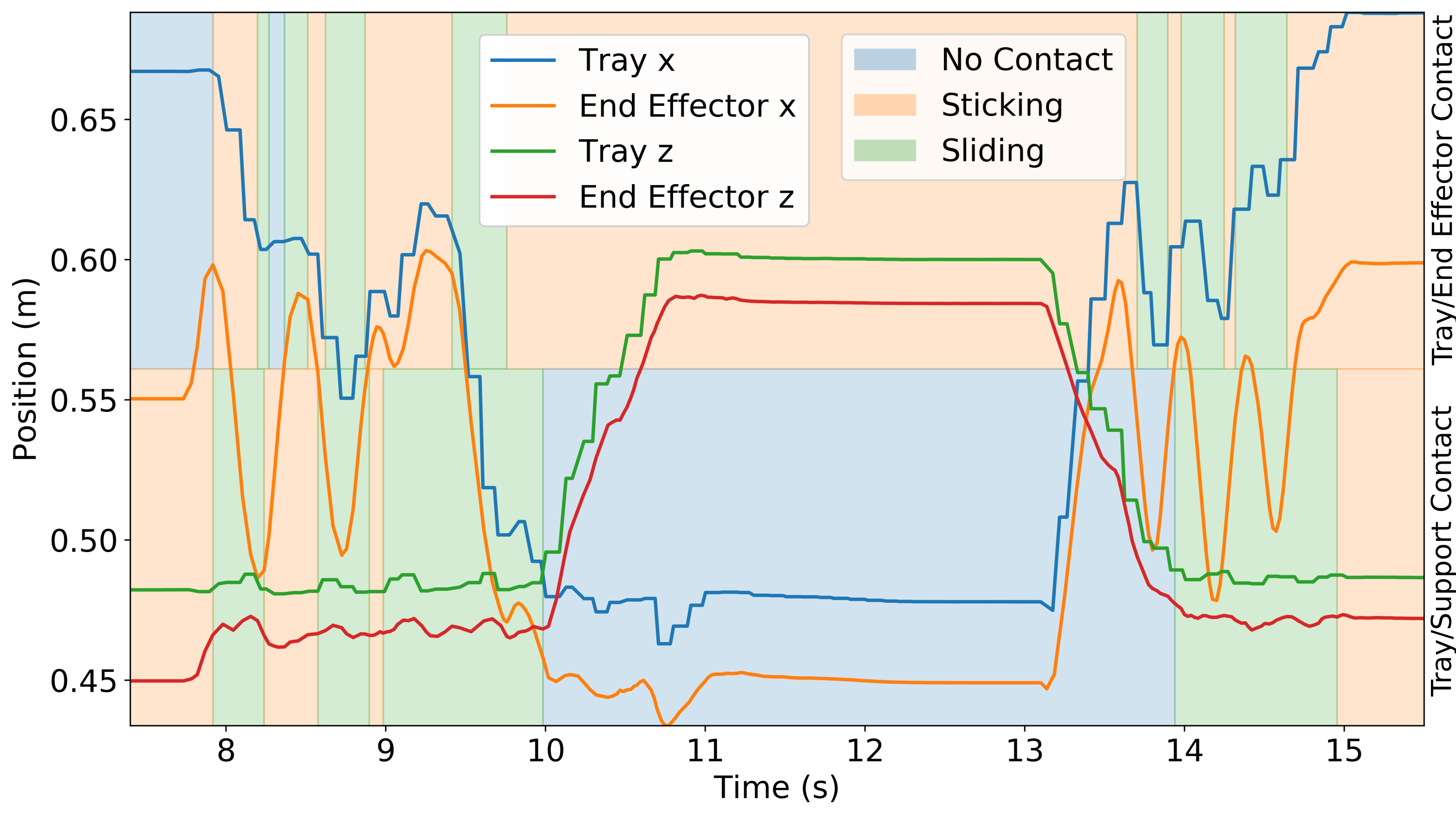}
	\caption{Position trajectories from execution on hardware. The visually estimated contact mode between the tray and end effector as well as tray and supports are indicated. Note, determining the actual contact mode for each of the seven contacts is challenging and there are likely many more contact mode transitions than reported in the figure. For instance, as shown in the last frame of \cref{fig:c3_plan}, each contact point can be active independently.}
	\label{fig:c3_actual_trajectory}
\end{figure}

First we ablate the contribution of tracking the end effector force by executing 10 experiments with and without the tracking objective.
The tracking controller with the end effector force objective succeeded for 80\% of the trials, failing once when trying to reach the second target when lifting an unbalanced tray and failing once to reach the third target when the tray slipped off in the direction of the robot base.
The trajectories of the end effector and tray for an execution are shown in \cref{fig:c3_actual_trajectory} and \cref{fig:trajectory_overlays}.
The tracking controller without the end effector force objective succeeded for 30\% of the trials, failing seven times to reach the third target due to the tray either colliding when the supports or slipping off in the direction of the robot base.

\subsection{Reliability Test}
The reliability of our method is evaluated by repeatedly executing the task without intervention.
This is possible without manual resetting because the final target position coincides with the initial position, and thus we treat tracking error from the previous execution as unstructured perturbations to the initial state.
Our method was able to complete six full cycles before failing due to the tray reach the position threshold (within 5cm) of the third target.

\subsection{Task Variations}
\label{subsec:task_variations}

\begin{figure}[h]
	\centering
	\begin{subfigure}[t]{0.235\textwidth}
		\includegraphics[width=1.0\textwidth]{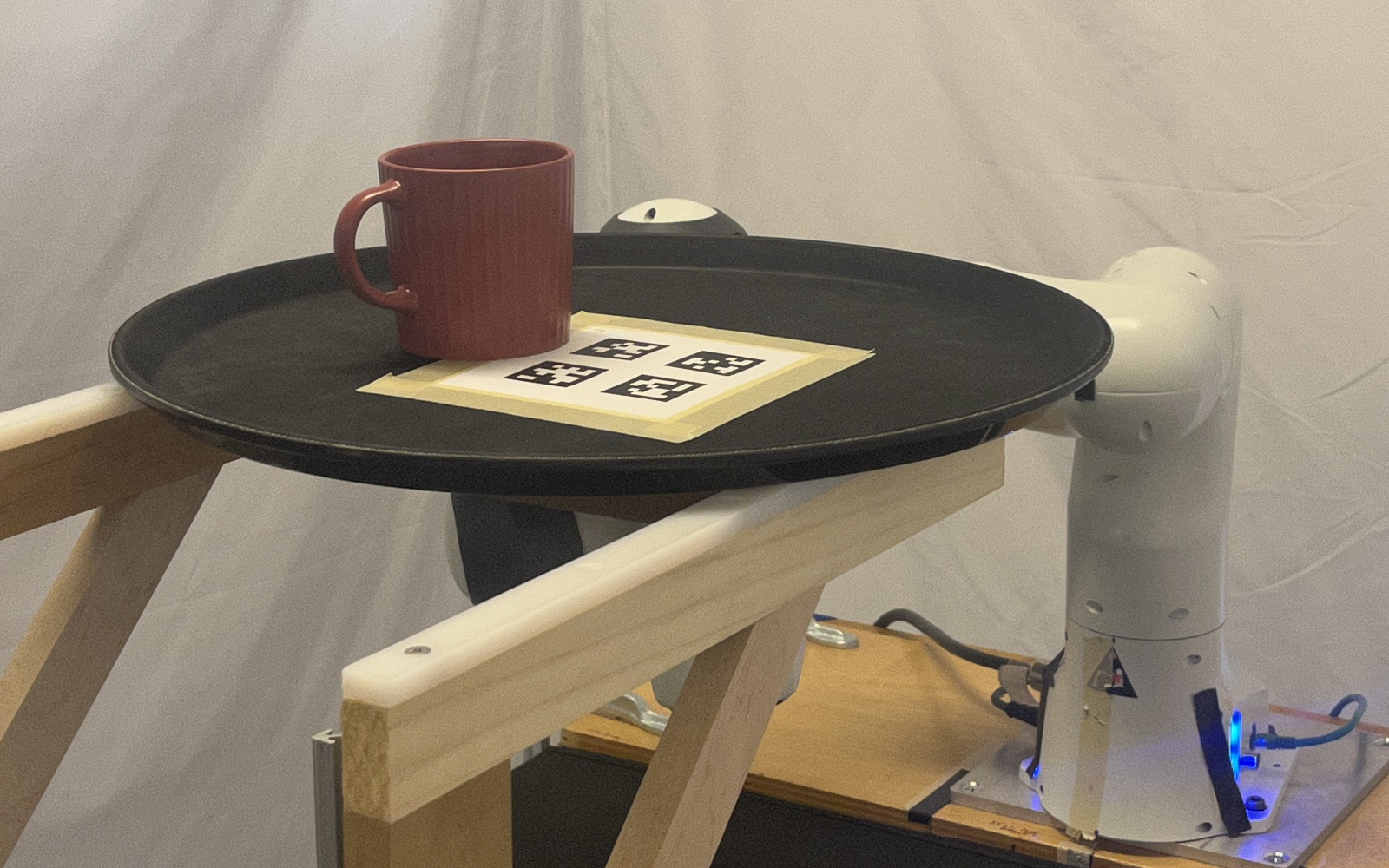}
		\caption{Unmodeled mug}
	\end{subfigure}	
	\begin{subfigure}[t]{0.235\textwidth}
		\includegraphics[width=1.0\textwidth]{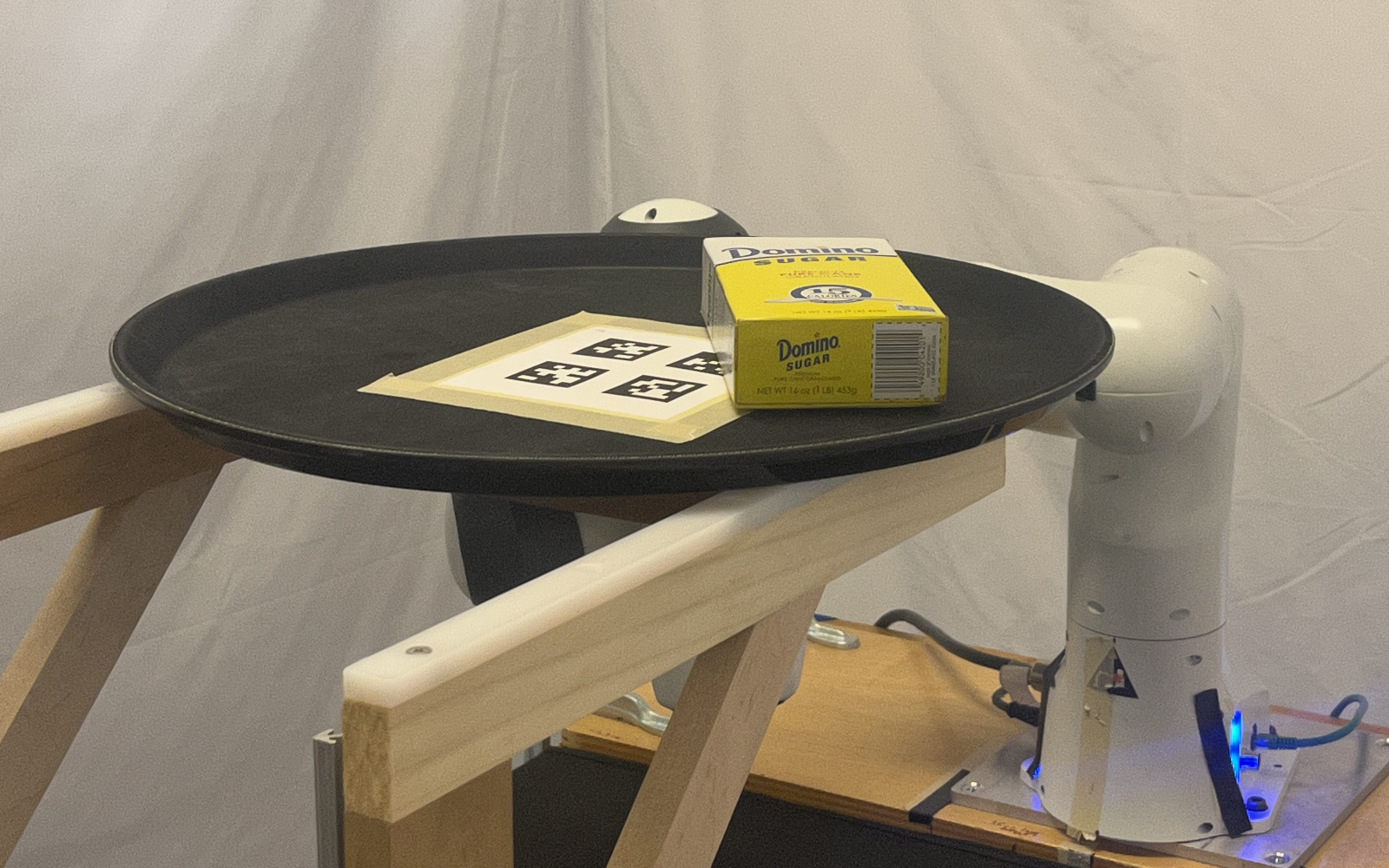}
		\caption{Unmodeled sugar box}
	\end{subfigure}	
	\begin{subfigure}[t]{0.48\textwidth}
		\includegraphics[width=1.0\textwidth]{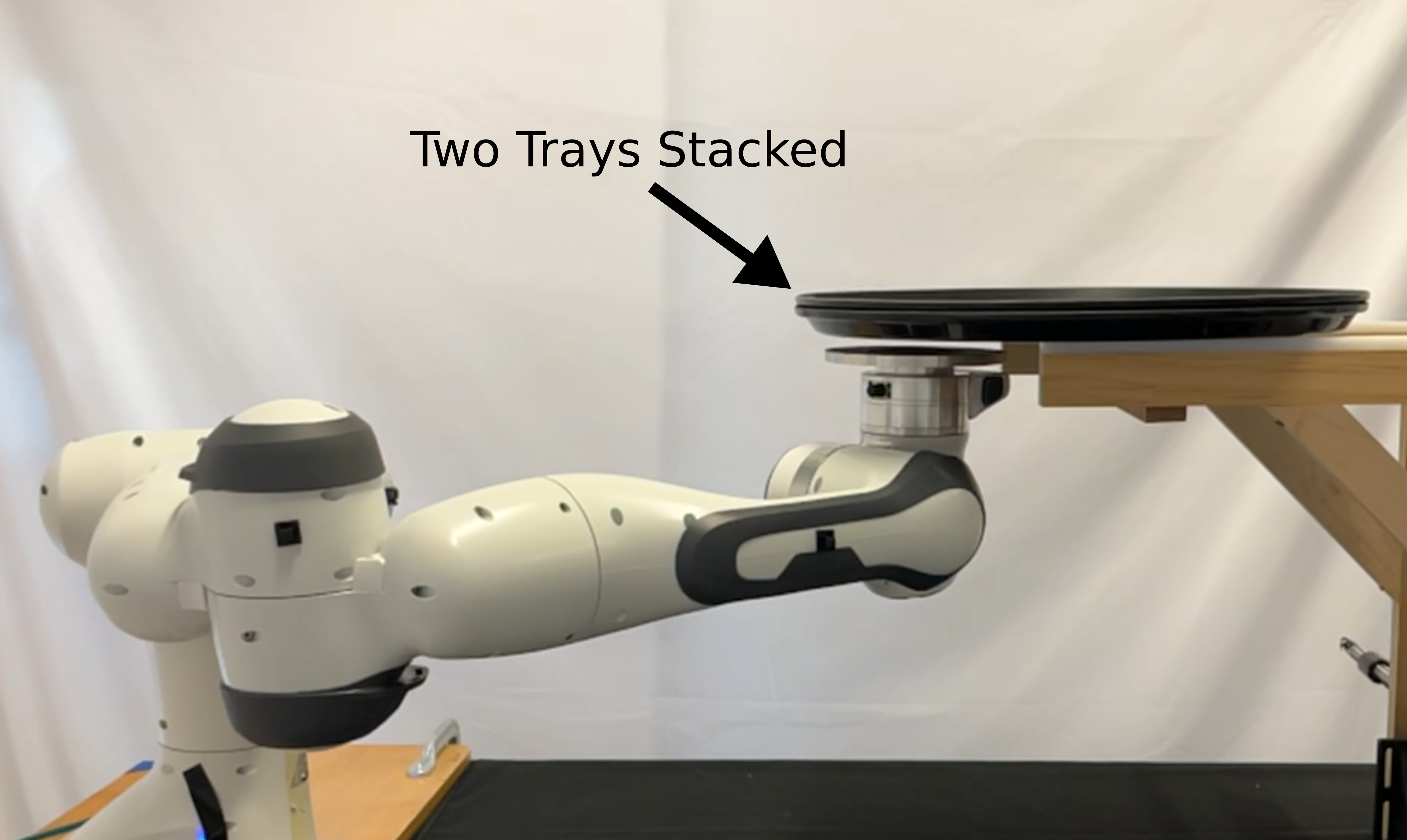}
		\caption{Two stacked trays (modeled in the controller)}
	\end{subfigure}	
	\caption{We evaluate our controller with the tray carrying unmodeled household objects placed at arbitrary positions as well as with two trays stacked on each other.}
	\label{fig:task_variations}
\end{figure}

To evaluate the robustness of our method to model error, specifically inaccurate mass and inertia properties, we add two different objects on top of the tray as shown in \cref{fig:task_variations}.
The first object is a common household mug that weighs 0.319 kg ($\sim$30\% mass of the tray).
We place it at an arbitrary position on the tray but take care to not obstruct the AprilTags on the tray.
We similarly test the tray with the second object, which is a sugar box that weighs 0.515 kg ($\sim$50\% mass of the tray).
Without adjusting any parameters, our controller is able to successfully complete the full task without failure.

Although our method demonstrates robustness to moderate model error, it fails when we double the mass by stacking two trays.
\sout{However, this can be easily addressed by updating the model of the tray used by our controller to reflect the increase in the mass and inertia values as well as the new contact geometry.}
\Revision{However, we can adjust our controller to accommodate the two stacked trays by updating its model to reflect its new mass, inertia, and contact geometry. Specifically, this means updating the corresponding values in the URDF file that defines the tray model.}
The task demonstrates the, perhaps obvious, finding that MPC is able leverage new object models.

\subsection{Behavior Analysis}
\label{subsec:friction_study}

\begin{figure}
	\begin{subfigure}[t]{0.48\textwidth}
		\includegraphics[width=1.0\textwidth]{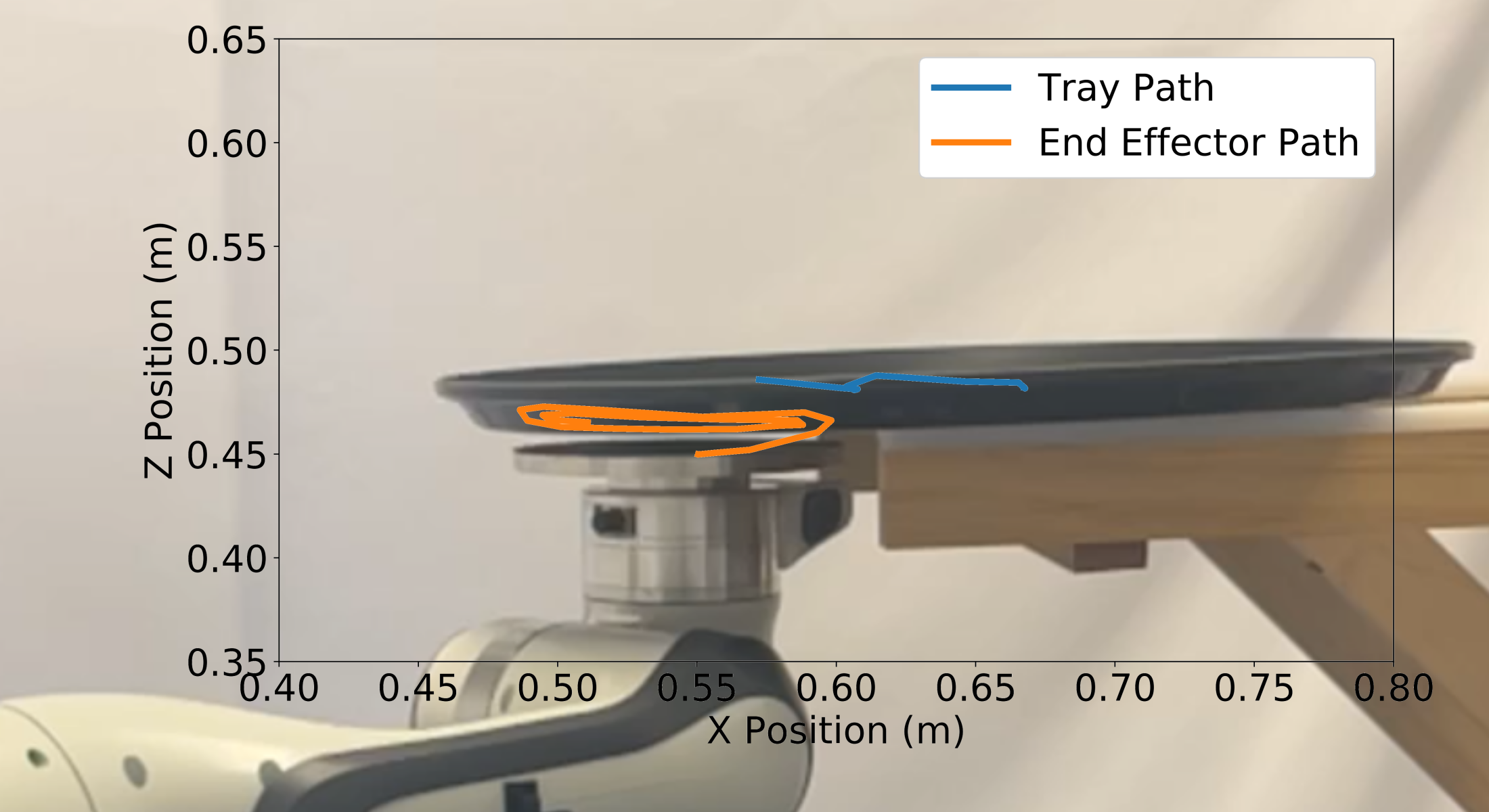}
		\caption{Retrieve Target}
	\end{subfigure}
	\begin{subfigure}[t]{0.48\textwidth}
		\includegraphics[width=1.0\textwidth]{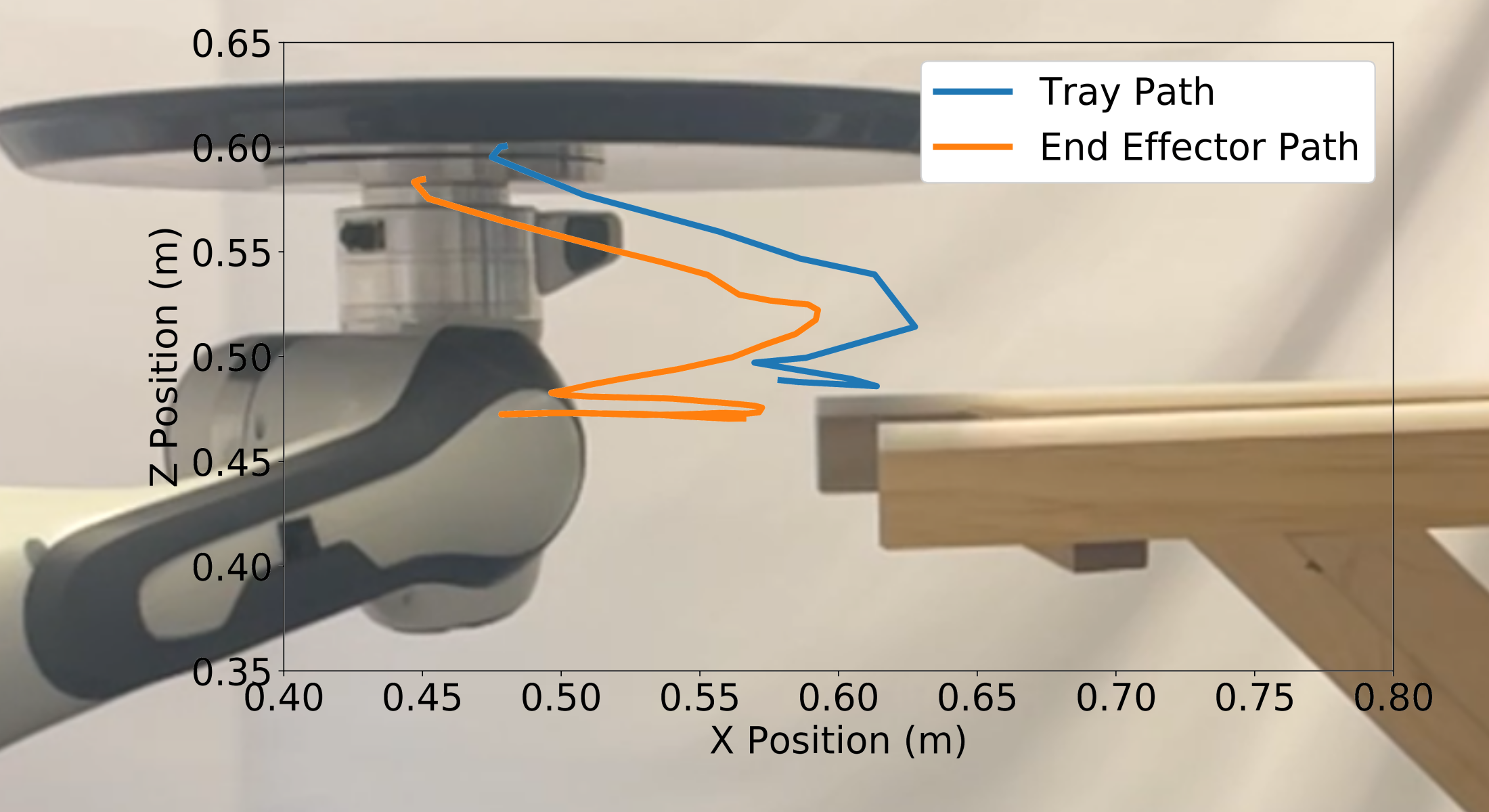}
		\caption{Place Target}
	\end{subfigure}	
	\caption{Portions of the end effector and tray trajectories approximately overlayed on top of image showing the naturally planned gaits from the MPC. The selected trajectory for the retrieve target (a) is 1.5 seconds long and 2.5 seconds long for the place target (b).}
	\label{fig:trajectory_overlays}
\end{figure}

We empirically observe that we did not need to tune any parameters, including friction, when transferring to hardware.
We hypothesize that controller feedback and the stick-slip ``gait" that naturally emerges from MPC has some inherent robustness to minor over \textit{and} under estimation of friction.
As evidence for this hypothesis, we observe the trajectory traces of the end effector and tray for two sections of the task where transitions between sliding and sticking contact are prevalent.
The first section is during the retrieval task when the controller attempts to slide the tray onto the end effector.
We plot a 1.5 second trajectory of the initial retrieval maneuver in \cref{fig:trajectory_overlays}, which shows that the end effector is not only moving back and forth along the direction of the target, but also raising and lowering in a circular pattern.
This gait increases the normal force between the tray when attempting to stick and decreases the normal force when attempting to slide, even utilizing the supports to entirely break contact with the tray.
This difference in contact forces results in a margin for the boundary between sticking and sliding.
The second section is during the place task when the controller attempts to slide the tray off of the end effector back onto the supports and is also shown in \cref{fig:trajectory_overlays}.
Here, the controller accelerates the tray forward and down in order to initiate sliding followed by a similar gait pattern as the first target once the tray is on the supports.
Although underestimating the friction force may not lead to failure as the tray should still reach the supports, overestimating the friction force may cause sliding during the initial forward acceleration.
This may explain why, during the ablation study, the most frequent failures were during this maneuver.

\Revision{
\subsection{Perturbation Recovery}
\label{subsec:perbutation_recovery}
The predominant motion of the task is along the x and z-axes.
To showcase the 3D nature of our method and to highlight its reactivity properties, we apply manual perturbations directly to the tray primarily along the y-axis during execution of the experiment.
Our controller is able to recover from modest perturbations applied during execution.
Footage of these perturbation recoveries are included in the supplemental video.
}

\begin{figure}
	\begin{subfigure}[h]{0.48\textwidth}
		\includegraphics[width=1.0\textwidth]{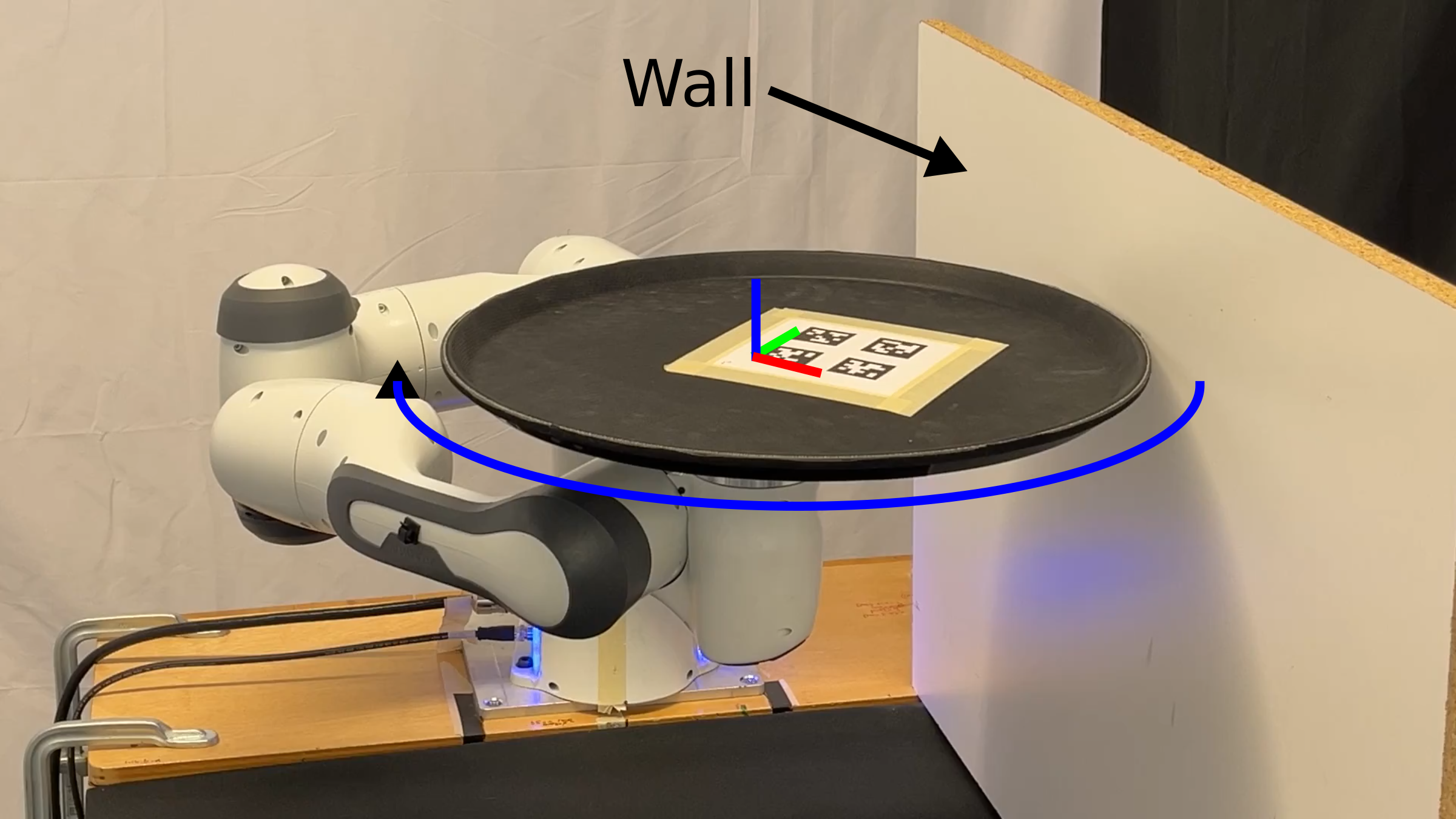}
		\caption{Initial position of the tray and desired rotation direction.}
	\end{subfigure}
	\begin{subfigure}[h]{0.48\textwidth}
		\includegraphics[width=1.0\textwidth]{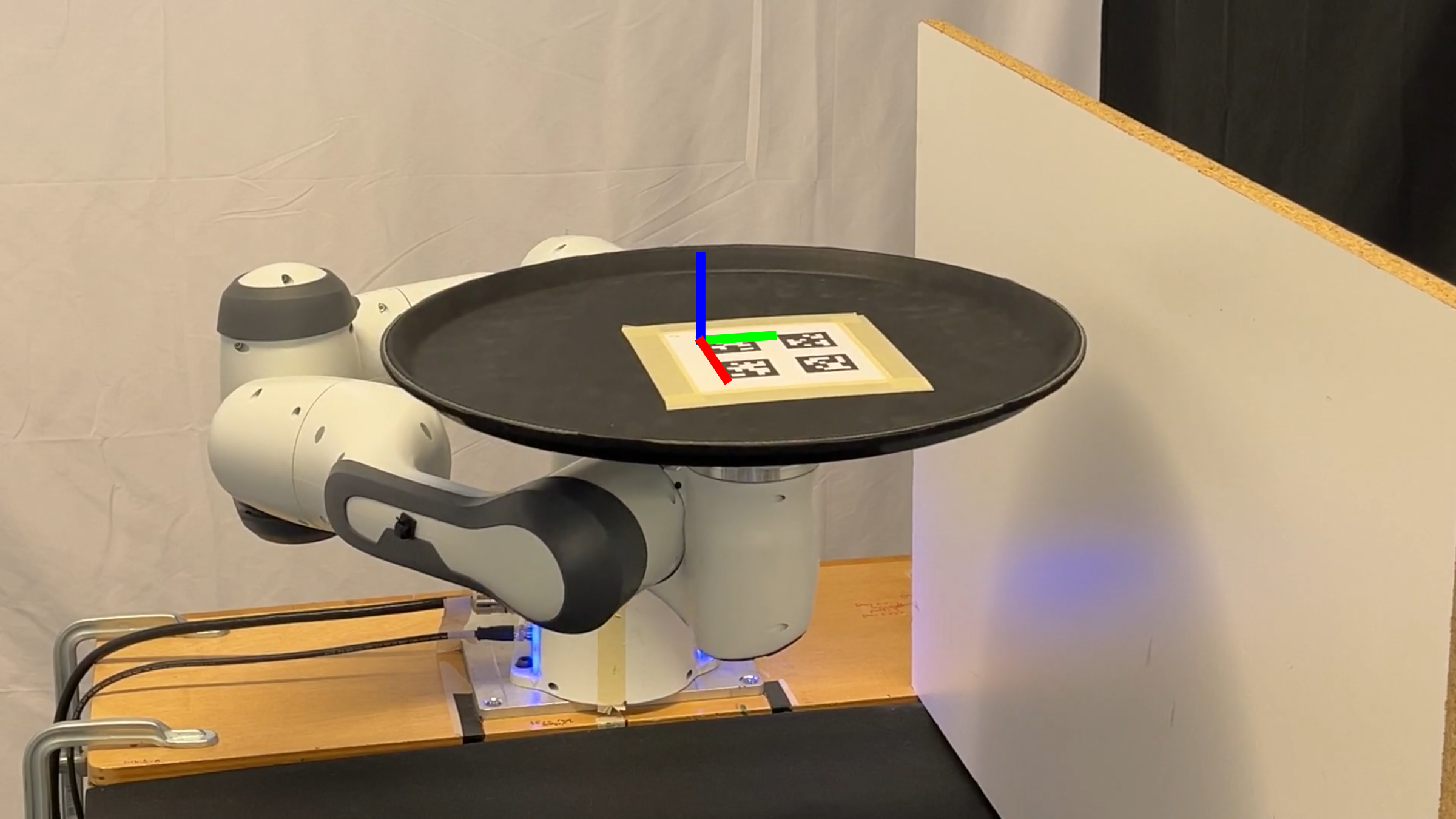}
		\caption{Final position of the tray after rotating using the wall.}
	\end{subfigure}
	\caption{We apply our framework to a different task where the robot is tasked with rotating the tray with the aid of a wall. Using the same LCS model for the end effector and tray and a single set of gains and a single target, our framework is able to successfully accomplish the task. The system is initialized (a) so that the tray must be rotated by approximately -45 degrees about the z-axis in order to reach the target configuration (bottom).}
	\label{fig:wall_task}
\end{figure}
\Revision{
\subsection{Tray Rotation using an External Wall}
To showcase the generality of our framework, we consider a different task where the robot is initialized with the tray balanced on the end effector and must rotate the tray using an external wall placed to the side of the robot as shown in \cref{fig:wall_task}.
Reorienting objects has many practical uses such as changing the viewing angle of the tray.
This task has similarities to previous works \cite{chavan2020planar} \cite{hou2018fast} that use external contacts to reorient an object grasped within parallel jaw grippers.
However, in our variation of the reorientation task, the object has no corners to use a pivot points.
In fact, only the tangential component from the external wall contact applies a useful moment for rotating the tray.
Additionally, because the tray is balanced on the end effector and not rigidly grasped, we have limited control authority to adjust the contact forces between the end effector and the tray.
The result is an underactuated reorientation task that requires careful planning and control of the sliding forces on the end effector and rolling between the tray and wall.
}

\Revision{
To specify this task, we are able to use the same representation (see \cref{subsec:full_c3_parameters}) for the end effector and tray in the LCS.
Therefore the only modeling change is that we replace the two supports with a single wall, which is represented as a simple box.
We show, still using a single set of MPC gains and a single target position for the tray and end effector, our framework successfully moves the tray to contact and wall and rotate the tray by 45 degrees around the z-axis in either direction.
Note, the gains used to perform this task are different from those used for the dynamic sliding task.
The gains, along with additional experiment details, are reported in \cref{subsec:appendix_external_wall}.
}
\section{Limitations}
\label{sec:limitations}

While our controller is fairly robust to mass and inertia, it is not robust to the height of the supports.
This is not surprising as the tray, supports, and robot are stiff, leading to senstivity at the boundary between contact and no-contact.
However, this can be addressed by quickly adapting the LCS parameters  \cite{huang2023adaptive} or generating the contact geometry and system dynamics entirely \cite{howell2022dojo} \cite{pfrommer2021contactnets}.

Another limitation of this method originates from the LCS model used by the MPC.
A known limitation of the LCS model is that it only is a linear representation of the contact decisions, meaning that it cannot consider contacts beyond the contact boundary.
For example, if the end effector moves entirely out from underneath the tray, the gap function $\phi$ will shift and instead consider only horizontal contact between the end effector and tray.
In this scenario, the MPC cannot find a solution for the task as it cannot reason about a path to go back under the tray.
This linear representation of contact boundaries can be limiting when applying this method to certain dynamic manipulation tasks such as flipping or scooping, however this can be addressed by linearizing about a reference trajectory with more informative configurations.
On the other hand, dynamic manipulation tasks such as batting and throwing should, in principle, fit well within our method.

While not needing to tune a separate set of parameters for each target highlights the flexibility of our approach, our selected parameters choose to favor broad motions over fine adjustments which results in higher steady state error.
An adaptive set of parameters depending on the task could reduce this tradeoff.
Additionally, the success of this task is extremely sensitive to the C3 parameters.
We found that many parameters need to be within 20\% of their final values.
Fortunately, the parameters that perform well in simulation also perform well on hardware as we did not adjust any parameters on hardware except to calibrate the height of the supports.

\Revision{Finally, the tasks we consider have relatively few contacts compared to the dexterous tasks demonstrated with robotic hands \cite{ma2023eureka}.
The number of contact variables scales linearly with the number of contacts and the possible MIQP branches per knot point scaling as $2^{n_{\lambda}}$. 
However, in practice, high performance can often be achieved without convergence of the ADMM iterations.
As a result, the compute time required to achieve sufficient performance is better than this worst-case analysis would imply.
However, this also means that the compute time is sensitive to the nuances in a given task, which might lead to faster or slower convergence to a high-quality policy.
We leave exploring the computational limits of MPC for future work.}

\Revision{\section{Conclusion}}
\label{sec:conclusion}

This paper demonstrates state-of-the-art performance for a on-palm sliding task, showing what can be accomplished with state feedback and contact-implicit MPC.

Although we show success with one particular formulation, there are many diverse formulations for contact-implicit MPC, which may have fundamentally different limitations or advantages.
Avenues for future research include evaluating the other formulations for contact-implicit MPC and understanding these fundamental differences.
Additionally, existing formulations seem to share common limitations, particularly the vulnerability to local minima.
Future work can address how to intelligently integrate higher-level planners into this framework.

\section*{Acknowledgments}

We would like to thank Alp Aydinoglu and Wei-Cheng Huang for helpful explanations about C3. We thank Bibit Bianchini and Mengti Sun for their tutorial on TagSlam. We thank Brian Acosta, Leon Kim, and Jessica Yin for their help with hardware experiments.

This material is based upon work supported by the
National Science Foundation Graduate Research Fellowship
Program under Grant No. DGE-1845298.
Additional funding was provided from the AI Institute.


\bibliographystyle{plainnat}
\bibliography{references}

\begin{thebibliography}{40}
\providecommand{\natexlab}[1]{#1}
\providecommand{\url}[1]{\texttt{#1}}
\expandafter\ifx\csname urlstyle\endcsname\relax
  \providecommand{\doi}[1]{doi: #1}\else
  \providecommand{\doi}{doi: \begingroup \urlstyle{rm}\Url}\fi

\bibitem[Anitescu and Potra(1997)]{anitescu1997formulating}
Mihai Anitescu and Florian~A Potra.
\newblock Formulating dynamic multi-rigid-body contact problems with friction
  as solvable linear complementarity problems.
\newblock \emph{Nonlinear Dynamics}, 14:\penalty0 231--247, 1997.
\newblock URL \url{https://link.springer.com/article/10.1023/A:1008292328909}.

\bibitem[Aydinoglu et~al.(2023)Aydinoglu, Wei, and
  Posa]{aydinoglu2023consensus}
Alp Aydinoglu, Adam Wei, and Michael Posa.
\newblock Consensus complementarity control for multi-contact mpc.
\newblock \emph{arXiv preprint arXiv:2304.11259}, 2023.
\newblock URL \url{https://arxiv.org/abs/2304.11259}.

\bibitem[Bledt(2020)]{bledt2020regularized}
Gerardo Bledt.
\newblock \emph{Regularized predictive control framework for robust dynamic
  legged locomotion}.
\newblock PhD thesis, Massachusetts Institute of Technology, 2020.
\newblock URL \url{https://dspace.mit.edu/handle/1721.1/125485}.

\bibitem[Brei et~al.(2024)Brei, Michaux, Zhang, Holmes, and
  Vasudevan]{brei2023serving}
Zachary Brei, Jonathan Michaux, Bohao Zhang, Patrick Holmes, and Ram Vasudevan.
\newblock Serving time: Real-time, safe motion planning and control for
  manipulation of unsecured objects.
\newblock \emph{IEEE Robotics and Automation Letters}, 9\penalty0 (3):\penalty0
  2383--2390, 2024.
\newblock \doi{10.1109/LRA.2024.3355731}.
\newblock URL \url{https://ieeexplore.ieee.org/document/10403905}.

\bibitem[Chavan-Dafle et~al.(2020)Chavan-Dafle, Holladay, and
  Rodriguez]{chavan2020planar}
Nikhil Chavan-Dafle, Rachel Holladay, and Alberto Rodriguez.
\newblock Planar in-hand manipulation via motion cones.
\newblock \emph{The International Journal of Robotics Research}, 39\penalty0
  (2-3):\penalty0 163--182, 2020.
\newblock URL
  \url{https://journals.sagepub.com/doi/full/10.1177/0278364919880257}.

\bibitem[Chen et~al.(2023)Chen, Tippur, Wu, Kumar, Adelson, and
  Agrawal]{chen2023visual}
Tao Chen, Megha Tippur, Siyang Wu, Vikash Kumar, Edward Adelson, and Pulkit
  Agrawal.
\newblock Visual dexterity: In-hand reorientation of novel and complex object
  shapes.
\newblock \emph{Science Robotics}, 8\penalty0 (84), 2023.
\newblock URL
  \url{https://www.science.org/doi/abs/10.1126/scirobotics.adc9244}.

\bibitem[Cheng et~al.(2022)Cheng, Huang, Hou, and Mason]{cheng2022contact}
Xianyi Cheng, Eric Huang, Yifan Hou, and Matthew~T. Mason.
\newblock Contact mode guided motion planning for quasidynamic dexterous
  manipulation in 3d.
\newblock In \emph{2022 International Conference on Robotics and Automation
  (ICRA)}, pages 2730--2736. IEEE, 2022.
\newblock URL \url{https://ieeexplore.ieee.org/document/9811872}.

\bibitem[Cheng et~al.(2023)Cheng, Patil, Temel, Kroemer, and
  Mason]{cheng2023enhancing}
Xianyi Cheng, Sarvesh Patil, Zeynep Temel, Oliver Kroemer, and Matthew~T Mason.
\newblock Enhancing dexterity in robotic manipulation via hierarchical contact
  exploration.
\newblock \emph{IEEE Robotics and Automation Letters}, 9\penalty0 (1):\penalty0
  390--397, 2023.
\newblock URL \url{https://ieeexplore.ieee.org/document/9811872}.

\bibitem[Chi et~al.(2023)Chi, Feng, Du, Xu, Cousineau, Burchfiel, and
  Song]{chi2023diffusion}
Cheng Chi, Siyuan Feng, Yilun Du, Zhenjia Xu, Eric Cousineau, Benjamin
  Burchfiel, and Shuran Song.
\newblock Diffusion policy: Visuomotor policy learning via action diffusion.
\newblock \emph{2023 Robotics Sciences and Systems}, 2023.
\newblock URL \url{https://roboticsconference.org/2023/program/papers/026/}.

\bibitem[Curtis et~al.(2022)Curtis, Fang, Kaelbling, Lozano-P{\'e}rez, and
  Garrett]{curtis2022long}
Aidan Curtis, Xiaolin Fang, Leslie~Pack Kaelbling, Tom{\'a}s Lozano-P{\'e}rez,
  and Caelan~Reed Garrett.
\newblock Long-horizon manipulation of unknown objects via task and motion
  planning with estimated affordances.
\newblock In \emph{2022 International Conference on Robotics and Automation
  (ICRA)}, pages 1940--1946. IEEE, 2022.
\newblock URL \url{https://ieeexplore.ieee.org/document/9812057}.

\bibitem[Doshi et~al.(2022)Doshi, Taylor, and Rodriguez]{doshi2022manipulation}
Neel Doshi, Orion Taylor, and Alberto Rodriguez.
\newblock Manipulation of unknown objects via contact configuration regulation.
\newblock In \emph{2022 International Conference on Robotics and Automation
  (ICRA)}, pages 2693--2699, 2022.
\newblock \doi{10.1109/ICRA46639.2022.9811713}.
\newblock URL \url{https://ieeexplore.ieee.org/document/9811713}.

\bibitem[Featherstone(2014)]{featherstone2014rigid}
Roy Featherstone.
\newblock \emph{Rigid body dynamics algorithms}.
\newblock Springer, 2014.
\newblock URL \url{https://link.springer.com/book/10.1007/978-1-4899-7560-7}.

\bibitem[Gilbert et~al.(1988)Gilbert, Johnson, and Keerthi]{gilbert1988fast}
Elmer~G Gilbert, Daniel~W Johnson, and S~Sathiya Keerthi.
\newblock A fast procedure for computing the distance between complex objects
  in three-dimensional space.
\newblock \emph{IEEE Journal on Robotics and Automation}, 4\penalty0
  (2):\penalty0 193--203, 1988.
\newblock URL \url{https://ieeexplore.ieee.org/document/2083}.

\bibitem[{Gurobi Optimization, LLC}(2023)]{gurobi}
{Gurobi Optimization, LLC}.
\newblock {Gurobi Optimizer Reference Manual}, 2023.
\newblock URL \url{https://www.gurobi.com}.

\bibitem[Heemels et~al.(2000)Heemels, Schumacher, and
  Weiland]{heemels2000linear}
WPMH Heemels, Johannes~M Schumacher, and S~Weiland.
\newblock Linear complementarity systems.
\newblock \emph{SIAM journal on applied mathematics}, 60\penalty0 (4):\penalty0
  1234--1269, 2000.
\newblock URL \url{https://doi.org/10.1137/S0036139997325199}.

\bibitem[Heins and Schoellig(2023)]{heins2023keep}
Adam Heins and Angela~P. Schoellig.
\newblock Keep it upright: Model predictive control for nonprehensile object
  transportation with obstacle avoidance on a mobile manipulator.
\newblock \emph{IEEE Robotics and Automation Letters}, 8\penalty0
  (12):\penalty0 7986--7993, 2023.
\newblock \doi{10.1109/LRA.2023.3324520}.
\newblock URL \url{https://ieeexplore.ieee.org/document/10285028}.

\bibitem[Higashimori et~al.(2009)Higashimori, Utsumi, Omoto, and
  Kaneko]{higashimori2009dynamic}
Mitsuru Higashimori, Keisuke Utsumi, Yasutaka Omoto, and Makoto Kaneko.
\newblock Dynamic manipulation inspired by the handling of a pizza peel.
\newblock \emph{IEEE Transactions on Robotics}, 25\penalty0 (4):\penalty0
  829--838, 2009.
\newblock \doi{10.1109/TRO.2009.2017085}.
\newblock URL \url{https://ieeexplore.ieee.org/document/4814586}.

\bibitem[Hou et~al.(2018)Hou, Jia, and Mason]{hou2018fast}
Yifan Hou, Zhenzhong Jia, and Matthew~T Mason.
\newblock Fast planning for 3d any-pose-reorienting using pivoting.
\newblock In \emph{2018 IEEE International Conference on Robotics and
  Automation (ICRA)}, pages 1631--1638. IEEE, 2018.
\newblock URL \url{https://ieeexplore.ieee.org/document/8462834}.

\bibitem[Hou et~al.(2020)Hou, Jia, Johnson, and Mason]{hou2020robust}
Yifan Hou, Zhenzhong Jia, Aaron~M Johnson, and Matthew~T Mason.
\newblock Robust planar dynamic pivoting by regulating inertial and grip
  forces.
\newblock In \emph{Algorithmic Foundations of Robotics XII: Proceedings of the
  Twelfth Workshop on the Algorithmic Foundations of Robotics}, pages 464--479.
  Springer, 2020.
\newblock URL
  \url{https://link.springer.com/chapter/10.1007/978-3-030-43089-4_30}.

\bibitem[Howell et~al.(2022)Howell, Cleac'h, Br{\"u}digam, Kolter, Schwager,
  and Manchester]{howell2022dojo}
Taylor~A Howell, Simon~Le Cleac'h, Jan Br{\"u}digam, J~Zico Kolter, Mac
  Schwager, and Zachary Manchester.
\newblock Dojo: A differentiable physics engine for robotics.
\newblock \emph{arXiv preprint arXiv:2203.00806}, 2022.
\newblock URL \url{https://arxiv.org/abs/2203.00806}.

\bibitem[Huang et~al.(2010)Huang, Olson, and Moore]{huang2010lcm}
Albert~S Huang, Edwin Olson, and David~C Moore.
\newblock Lcm: Lightweight communications and marshalling.
\newblock In \emph{2010 IEEE/RSJ International Conference on Intelligent Robots
  and Systems}, pages 4057--4062. IEEE, 2010.
\newblock URL \url{https://ieeexplore.ieee.org/document/5649358}.

\bibitem[Huang et~al.(2023)Huang, Aydinoglu, Jin, and Posa]{huang2023adaptive}
Wei-Cheng Huang, Alp Aydinoglu, Wanxin Jin, and Michael Posa.
\newblock Adaptive contact-implicit model predictive control with online
  residual learning.
\newblock \emph{arXiv preprint arXiv:2310.09893}, 2023.
\newblock URL \url{https://arxiv.org/abs/2310.09893}.

\bibitem[Khatib(1987)]{khatib1987unified}
O.~Khatib.
\newblock A unified approach for motion and force control of robot
  manipulators: The operational space formulation.
\newblock \emph{IEEE Journal on Robotics and Automation}, 3\penalty0
  (1):\penalty0 43--53, 1987.
\newblock \doi{10.1109/JRA.1987.1087068}.
\newblock URL \url{https://ieeexplore.ieee.org/document/1087068}.

\bibitem[Kurtz et~al.(2023)Kurtz, Castro, {\"O}nol, and Lin]{kurtz2023inverse}
Vince Kurtz, Alejandro Castro, Aykut~{\"O}zg{\"u}n {\"O}nol, and Hai Lin.
\newblock Inverse dynamics trajectory optimization for contact-implicit model
  predictive control.
\newblock \emph{arXiv preprint arXiv:2309.01813}, 2023.
\newblock URL \url{https://arxiv.org/abs/2309.01813}.

\bibitem[Le~Cleac'h et~al.(2024)Le~Cleac'h, Howell, Yang, Lee, Zhang, Bishop,
  Schwager, and Manchester]{le2024fast}
Simon Le~Cleac'h, Taylor~A. Howell, Shuo Yang, Chi-Yen Lee, John Zhang, Arun
  Bishop, Mac Schwager, and Zachary Manchester.
\newblock Fast contact-implicit model predictive control.
\newblock \emph{IEEE Transactions on Robotics}, pages 1--14, 2024.
\newblock \doi{10.1109/TRO.2024.3351554}.
\newblock URL \url{https://ieeexplore.ieee.org/document/10384795}.

\bibitem[Lynch and Mason(1999)]{lynch1999dynamic}
Kevin~M Lynch and Matthew~T Mason.
\newblock Dynamic nonprehensile manipulation: Controllability, planning, and
  experiments.
\newblock \emph{The International Journal of Robotics Research}, 18\penalty0
  (1):\penalty0 64--92, 1999.
\newblock URL
  \url{https://journals.sagepub.com/doi/abs/10.1177/027836499901800105}.

\bibitem[Ma et~al.(2023)Ma, Liang, Wang, Huang, Bastani, Jayaraman, Zhu, Fan,
  and Anandkumar]{ma2023eureka}
Yecheng~Jason Ma, William Liang, Guanzhi Wang, De-An Huang, Osbert Bastani,
  Dinesh Jayaraman, Yuke Zhu, Linxi Fan, and Anima Anandkumar.
\newblock Eureka: Human-level reward design via coding large language models.
\newblock \emph{arXiv preprint arXiv:2310.12931}, 2023.

\bibitem[Masterjohn et~al.(2022)Masterjohn, Guoy, Shepherd, and
  Castro]{masterjohn2022velocity}
Joseph Masterjohn, Damrong Guoy, John Shepherd, and Alejandro Castro.
\newblock Velocity level approximation of pressure field contact patches.
\newblock \emph{IEEE Robotics and Automation Letters}, 7\penalty0 (4):\penalty0
  11593--11600, 2022.
\newblock \doi{10.1109/LRA.2022.3203845}.
\newblock URL \url{https://ieeexplore.ieee.org/document/9874987}.

\bibitem[Pfrommer and Daniilidis(2019)]{pfrommer2019tagslam}
Bernd Pfrommer and Kostas Daniilidis.
\newblock Tagslam: Robust slam with fiducial markers.
\newblock \emph{arXiv preprint arXiv:1910.00679}, 2019.
\newblock URL \url{https://github.com/berndpfrommer/tagslam_root}.

\bibitem[Pfrommer et~al.(2021)Pfrommer, Halm, and
  Posa]{pfrommer2021contactnets}
Samuel Pfrommer, Mathew Halm, and Michael Posa.
\newblock Contactnets: Learning discontinuous contact dynamics with smooth,
  implicit representations.
\newblock In Jens Kober, Fabio Ramos, and Claire Tomlin, editors,
  \emph{Proceedings of the 2020 Conference on Robot Learning}, volume 155 of
  \emph{Proceedings of Machine Learning Research}, pages 2279--2291. PMLR,
  16--18 Nov 2021.
\newblock URL \url{https://proceedings.mlr.press/v155/pfrommer21a.html}.

\bibitem[Pham et~al.(2017)Pham, Caron, Lertkultanon, and
  Nakamura]{pham2017admissible}
Quang-Cuong Pham, St{\'e}phane Caron, Puttichai Lertkultanon, and Yoshihiko
  Nakamura.
\newblock Admissible velocity propagation: Beyond quasi-static path planning
  for high-dimensional robots.
\newblock \emph{The International Journal of Robotics Research}, 36\penalty0
  (1):\penalty0 44--67, 2017.
\newblock \doi{10.1177/0278364916675419}.
\newblock URL \url{https://doi.org/10.1177/0278364916675419}.

\bibitem[Remy(2017)]{remy2017ambiguous}
C~David Remy.
\newblock Ambiguous collision outcomes and sliding with infinite friction in
  models of legged systems.
\newblock \emph{The International Journal of Robotics Research}, 36\penalty0
  (12):\penalty0 1252--1267, 2017.
\newblock URL \url{https://doi.org/10.1177/0278364917731820}.

\bibitem[Ruggiero et~al.(2018)Ruggiero, Lippiello, and
  Siciliano]{ruggiero2018nonprehensile}
Fabio Ruggiero, Vincenzo Lippiello, and Bruno Siciliano.
\newblock Nonprehensile dynamic manipulation: A survey.
\newblock \emph{IEEE Robotics and Automation Letters}, 3\penalty0 (3):\penalty0
  1711--1718, 2018.
\newblock \doi{10.1109/LRA.2018.2801939}.
\newblock URL \url{https://ieeexplore.ieee.org/document/8280543}.

\bibitem[Shi et~al.(2017)Shi, Woodruff, Umbanhowar, and Lynch]{shi2017dynamic}
Jian Shi, J~Zachary Woodruff, Paul~B Umbanhowar, and Kevin~M Lynch.
\newblock Dynamic in-hand sliding manipulation.
\newblock \emph{IEEE Transactions on Robotics}, 33\penalty0 (4):\penalty0
  778--795, 2017.
\newblock URL \url{https://ieeexplore.ieee.org/document/7913727}.

\bibitem[Stellato et~al.(2020)Stellato, Banjac, Goulart, Bemporad, and
  Boyd]{stellato2020osqp}
Bartolomeo Stellato, Goran Banjac, Paul Goulart, Alberto Bemporad, and Stephen
  Boyd.
\newblock Osqp: An operator splitting solver for quadratic programs.
\newblock \emph{Mathematical Programming Computation}, 12\penalty0
  (4):\penalty0 637--672, 2020.
\newblock URL
  \url{https://link.springer.com/article/10.1007/s12532-020-00179-2}.

\bibitem[Subburaman et~al.(2023)Subburaman, Selvaggio, and
  Ruggiero]{subburaman2023non}
Rajesh Subburaman, Mario Selvaggio, and Fabio Ruggiero.
\newblock A non-prehensile object transportation framework with adaptive
  tilting based on quadratic programming.
\newblock \emph{IEEE Robotics and Automation Letters}, 2023.
\newblock URL \url{https://ieeexplore.ieee.org/document/10105969}.

\bibitem[Taylor et~al.(2023)Taylor, Doshi, and Rodriguez]{taylor2023object}
Orion Taylor, Neel Doshi, and Alberto Rodriguez.
\newblock Object manipulation through contact configuration regulation:
  Multiple and intermittent contacts.
\newblock In \emph{2023 IEEE/RSJ International Conference on Intelligent Robots
  and Systems (IROS)}, pages 8735--8743, 2023.
\newblock \doi{10.1109/IROS55552.2023.10341362}.
\newblock URL \url{https://ieeexplore.ieee.org/document/10341362}.

\bibitem[Tedrake and the Drake Development~Team(2019)]{drake}
Russ Tedrake and the Drake Development~Team.
\newblock Drake: Model-based design and verification for robotics, 2019.
\newblock URL \url{https://drake.mit.edu}.

\bibitem[Wensing and Orin(2013)]{wensing2013generation}
Patrick~M. Wensing and David~E. Orin.
\newblock Generation of dynamic humanoid behaviors through task-space control
  with conic optimization.
\newblock In \emph{2013 IEEE International Conference on Robotics and
  Automation}, pages 3103--3109, 2013.
\newblock \doi{10.1109/ICRA.2013.6631008}.
\newblock URL \url{https://ieeexplore.ieee.org/document/6631008}.

\bibitem[Woodruff and Lynch(2017)]{woodruff2017planning}
J.~Zachary Woodruff and Kevin~M. Lynch.
\newblock Planning and control for dynamic, nonprehensile, and hybrid
  manipulation tasks.
\newblock In \emph{2017 IEEE International Conference on Robotics and
  Automation (ICRA)}, pages 4066--4073, 2017.
\newblock \doi{10.1109/ICRA.2017.7989467}.
\newblock URL \url{https://ieeexplore.ieee.org/document/7989467/}.

\end{thebibliography}

\newpage

\section{Appendices}
\label{sec:appendix}

\subsection{\Revision{Full C3 Parameters for Dynamic Sliding Task}}
\label{subsec:full_c3_parameters}

We report the full C3 parameters used across all experiments.
We refer to \citet{aydinoglu2023consensus} for the detailed parameter definitions.
The LCS state vector for our system is $x_{lcs} = [q_{lcs}, v_{lcs}]$, where 
$$q_{lcs} = 
\begin{bmatrix}
ee_{x}\\
ee_{y}\\
ee_{z}\\
tray_{qw}\\
tray_{qx}\\
tray_{qy}\\
tray_{qz}\\
tray_{x}\\
tray_{y}\\
tray_{z}\end{bmatrix},
v_{lcs} = 
\begin{bmatrix}
ee_{vx}\\
ee_{vy}\\
ee_{vz}\\
tray_{wx}\\
tray_{wy}\\
tray_{wz}\\
tray_{vx}\\
tray_{vy}\\
tray_{vz}\end{bmatrix}.$$
All quantities are expressed in the world frame.
$()_x$ indicates the x position, $()_{qw, qx, qy, qz}$ is the orientation expressed as a quaternion, $()_{vy}$ indicates the y velocity, and $()_{wz}$ expressed the angular velocity.
The LCS input vector is $u_{lcs} = [u_x, u_y, u_z]$ expressed as forces applied to the end effector.
The contact forces are $\lambda \in \Real^{4 n_{contacts}}$, where $n_{contacts}$ is 7 for our problem. Reminder that 4 comes from the 4 extreme rays of a pyramidal approximation of the friction cone.
We report the parameters in \cref{tab:full_c3_parameters}.
The matrices $Q, R, G, U$ are all diagonal matrices, so we report the diagonal terms for conciseness.
We use only three values to parameterize $G$ and $U$, once each for the state variables, contact variables, and input variables, where $G$ and $U$ are diagonal matrices constructed from three diagonal matrices as 
$G = \begin{bmatrix} G_x & & \\
& G_{\lambda} & \\
& & G_{u} \end{bmatrix}$,
and $G_x = w_{G_x} I$, where $G_x$ is overloaded to mean both diagonal matrix and the scalar that defines the matrix.
Workspace limits are imposed only on the end effector as $q_{ee, min} \leq q_{ee} \leq q_{ee, max}$.

\begin{table}[h]
	\centering
	\resizebox{0.48\textwidth}{!}{
	\begin{tabular}{|c|c|}
		\hline
		$N$ & 5\\
		$dt$ & 0.075\\
		$\mu_{tray, ee}$ & 0.6 \\
		$\mu_{tray, supports}$ & 0.1\\
		$\rho$ & 4\\
		ADMM iterations & 2\\
		$Q_q$ & 50 * [150, 150, 150, 0, 1, 1, 0, 15000, 15000, 15000]\\
		$Q_v$ & 50 * [5, 5, 15, 10, 10, 1, 5, 5, 5]\\
		$R$ & 50 * [0.15, 0.15, 0.1]\\
		$w_{G_x}$ & 0.1 \\
		$w_{G_{\lambda}}$ & 10\\
		$w_{G_{u}}$ & 0.1\\
		$w_{U_x}$ & 0.1 \\
		$w_{U_{\lambda}}$ & 10\\
		$w_{U_{u}}$ & 3\\
		$u_{min}$ & [-10, -10, 0] \\
		$u_{max}$ & [10, 10, 30] \\
		$q_{ee, min}$ & [0.4, -0.1, 0.35]\\
		$q_{ee, max}$ & [0.6, 0.1, 0.7]\\
		\hline
	\end{tabular}
	}
	\caption{Full C3 parameters used across all tray retrieval experiments}
	\label{tab:full_c3_parameters}
\end{table}
%
%
\Revision{
\subsection{Additional Details for Rotating with an External Wall}
\label{subsec:appendix_external_wall}
We use the same LCS model to represent the state of the end effector and tray as the tray retrieval task.
Thus, $x_{lcs}$ and $u_{lcs}$ for this task are the same as discussed in \cref{subsec:full_c3_parameters}.
However, for this task, $n_{contacts}$ is 4.
This includes the 3 contacts to model the surface-surface contact between the end effector and the tray and 1 contact to model the interaction between the tray and the wall.
The MPC gains used for this task are reported below in \cref{tab:full_c3_parameters_wall}.
}
\Revision{
We give just a single fixed target $$q_{lcs, des} = [0.55, 0.0, 0.469, 1, 0, 0, 0, 0.55, 0.0, 0.485],$$ for this task and the state of the system is initialized to approximately
$$q_{lcs, init} = [0.55, 0.0, 0.469, 0.925, 0, 0, 0.38, 0.55, 0.02, 0.485].$$
The quaternion $[0.925, 0, 0, 0.38]$ is approximately a rotation of 45 degrees around its z-axis.
We also successfully perform same experiment with the tray rotated approximately by -45 degrees around its z-axis.
We found that offsetting the position of tray in the direction of the wall encouraged the MPC to utilize the wall, because otherwise it would have to trade off position error with orientation error instead of reducing both simultaneously.
Additionally, the wall is placed 0.3 m to the side of the robot in order to decrease the penalty of using the wall, as the robot would need to move away from the target in order to make contact with the wall.
Finally, because the state tracking terms of the MPC error $(x_{lcs,des} - x_{lcs})^T Q (x_{lcs,des} - x_{lcs})$ is improperly defined for the quaternions components, we convert the quaternion orientation error into angle-axis form and set the desired tray angular velocity to be proportional to that error.}

\Revision{
In the hardware setup, we use a single particle board as the wall and use the same tray as the tray retrieval task.
Additionally, we add high friction tape to the rim of the tray in order to increase the friction of that surface.
}
\begin{table}[h]
	\centering
	\resizebox{0.48\textwidth}{!}{
	\begin{tabular}{|c|c|}
		\hline
		$N$ & 4\\
		$dt$ & 0.05\\
		$\mu_{tray, ee}$ & 0.8 \\
		$\mu_{tray, wall}$ & 1.0\\
		$\rho$ & 5\\
		ADMM iterations & 3\\
		$Q_q$ & 50 * [10, 10, 150, 1000, 1000, 1000, 1000, 25, 25, 15000]\\
		$Q_v$ & 50 * [5, 5, 5, 1, 1, 500, 5, 5, 5]\\
		$R$ & 75 * [1.9, 0.5, 0.05]\\
		$w_{G_x}$ & 0.5 \\
		$w_{G_{\lambda}}$ & 75\\
		$w_{G_{u}}$ & 1.25\\
		$w_{U_x}$ & 0.5 \\
		$w_{U_{\lambda}}$ & 50\\
		$w_{U_{u}}$ & 15\\
		$u_{min}$ & [-10, -10, 0] \\
		$u_{max}$ & [10, 10, 30] \\
		$q_{ee, min}$ & [0.45, -0.2, 0.4]\\
		$q_{ee, max}$ & [0.7, 0.2, 0.5]\\
		\hline
	\end{tabular}
	}
	\caption{Full C3 parameters used for rotating with external wall experiment}
	\label{tab:full_c3_parameters_wall}
\end{table}

\end{document}